\documentclass[conference]{IEEEtran}
\usepackage[latin9]{inputenc}
\usepackage{amsmath}
\usepackage{graphicx}
\usepackage{amssymb}
\usepackage{algorithm}
\usepackage[noend]{algpseudocode}
\usepackage{url}

\def\eg{\emph{e.g.}}

\def\ie{\emph{i.e.}}

\def\I{\mathbf{I}}

\def\H{\mathbf{H}}

\def\F{\mathbf{F}}

\ifCLASSINFOpdf
\else
\fi

\usepackage{stfloats}

\usepackage{multirow}
\renewcommand{\sl}[1]{\textcolor{red}{[SL: #1]}}

\usepackage{color}

\IEEEoverridecommandlockouts

\begin{document}

%
% paper title
% Titles are generally capitalized except for words such as a, an, and, as,
% at, but, by, for, in, nor, of, on, or, the, to and up, which are usually
% not capitalized unless they are the first or last word of the title.
% Linebreaks \\ can be used within to get better formatting as desired.
% Do not put math or special symbols in the title.
\title{Landmark Breaker: Obstructing DeepFake By Disturbing Landmark Extraction}

% author names and affiliations
% use a multiple column layout for up to three different
% affiliations
% Yo
\author{Pu Sun$^{\dagger*}$, Yuezun Li$^{\ddagger*}$\thanks{$^*$ The authors contribute equally}, Honggang Qi$^\dagger$ and Siwei Lyu$^\ddagger$  \\
$^\dagger$ University of Chinese Academy of Sciences, China \\
$^\ddagger$ University at Buffalo, State University of New York, USA}

% conference papers do not typically use \thanks and this command
% is locked out in conference mode. If really needed, such as for
% the acknowledgment of grants, issue a \IEEEoverridecommandlockouts
% after \documentclass

% for over three affiliations, or if they all won't fit within the width
% of the page, use this alternative format:
%
%\author{\IEEEauthorblockN{Michael Shell\IEEEauthorrefmark{1},
%Homer Simpson\IEEEauthorrefmark{2},
%James Kirk\IEEEauthorrefmark{3},
%Montgomery Scott\IEEEauthorrefmark{3} and
%Eldon Tyrell\IEEEauthorrefmark{4}}
%\IEEEauthorblockA{\IEEEauthorrefmark{1}School of Electrical and Computer Engineering\\
%Georgia Institute of Technology,
%Atlanta, Georgia 30332--0250\\ Email: see http://www.michaelshell.org/contact.html}
%\IEEEauthorblockA{\IEEEauthorrefmark{2}Twentieth Century Fox, Springfield, USA\\
%Email: homer@thesimpsons.com}
%\IEEEauthorblockA{\IEEEauthorrefmark{3}Starfleet Academy, San Francisco, California 96678-2391\\
%Telephone: (800) 555--1212, Fax: (888) 555--1212}
%\IEEEauthorblockA{\IEEEauthorrefmark{4}Tyrell Inc., 123 Replicant Street, Los Angeles, California 90210--4321}}

% use for special paper notices
%\IEEEspecialpapernotice{(Invited Paper)}

% make the title area
\maketitle

% %INCLUDES COPYRIGHT NOTICE: one of three copyright notice should be included.
% %Uncomment the appropriate line below, according to the authors %affiliation:
% \begin{figure}[b]
% \vspace{-0.3cm}
% \parbox{\hsize}{\em
% %information about the event:
% WIFS`2020, December, 6-9, 2020, New York, USA.
% %copyright notice: one of four copyright notices below should be included. Choose the right one below according to the authors affiliation:
% %XXX-X-XXXX-XXXX-X/XX/\$XX.00 \ \copyright 2017 European Union.
% %XXX-X-XXXX-XXXX-X/XX/\$XX.00  \ \copyright 2017 Crown.
% %U.S. Government work not protected by U.S. copyright.
% %XXX-X-XXXX-XXXX-X/XX/\$XX.00 \ \copyright 2017 IEEE.
% XXX-X-XXXX-XXXX-X/XX/\$XX.00 \ \copyright 2020 IEEE.
% }\end{figure}

%INCLUDES COPYRIGHT NOTICE: one of three copyright notice should be included.

%Uncomment the appropriate line below, according to the authors %affiliation:

\begin{figure}[b]

\vspace{-0.3cm}

\parbox{\hsize}{\em

%information about the event:

WIFS`2020, December, 6-11, 2020, New York, USA.

%copyright notice: one of four copyright notices below should be included. Choose the right one below according to the authors affiliation:

% 978-1-7281-9930-6/20/\$31.00 \ \copyright 2017 European Union.

% 978-1-7281-9930-6/20/\$31.00  \ \copyright 2017 Crown.

% U.S. Government work not protected by U.S. copyright.

978-1-7281-9930-6/20/\$31.00 \ \copyright 2020 IEEE.

}\end{figure}

% As a general rule, do not put math, special symbols or citations
% in the abstract
\begin{abstract}
The recent development of Deep Neural Networks (DNN) has significantly increased the realism of AI-synthesized faces, with the most notable examples being the DeepFakes. The DeepFake technology can synthesize a face of target subject from a face of another subject, while retains the same face attributes. With the rapidly increased social media portals (Facebook, Instagram, etc), these realistic fake faces rapidly spread though the Internet, causing a broad negative impact to the society. 
In this paper, we describe {\em Landmark Breaker}, the first dedicated method to disrupt facial landmark extraction, and apply it to the obstruction of the generation of DeepFake videos. 
% \sl{add one sentence about how LB works, and another sentence about how it is used to obstruct DF generation.} 
Our motivation is that disrupting the facial landmark extraction can affect the alignment of input face so as to degrade the DeepFake quality. Our method is achieved using adversarial perturbations.
Compared to the detection methods that only work after DeepFake generation, Landmark Breaker goes one step ahead to prevent DeepFake generation. The experiments are conducted on three state-of-the-art facial landmark extractors using the recent Celeb-DF dataset.
\end{abstract}

% no keywords

% For peer review papers, you can put extra information on the cover
% page as needed:
% \ifCLASSOPTIONpeerreview
% \begin{center} \bfseries EDICS Category: 3-BBND \end{center}
% \fi
%
% For peerreview papers, this IEEEtran command inserts a page break and
% creates the second title. It will be ignored for other modes.
\IEEEpeerreviewmaketitle

\section{Introduction}

% The privacy protection has been drawn increasingly attention due to the tremendous personal data (\eg, selfie image or videos) circulated publicly through social media such as YouTube, FaceBook and Instagram. 
% These vast volume of personal data exacerbates the potential risk of privacy leakage, as they can be used unauthorized in privacy related applications such as face recognition or head pose estimation.

% Recently, the demand of privacy protection becomes more urgent due to the drastically improvement of deep neural networks on face synthesis, especially the advent of technology named DeepFake. With the large amount of personal data, DeepFake can easily synthesize a realistic face of target individual based on a real face of donor individual, while keeps the same facial expression and orientation. Therefore, it can create illusions of a person's presence and activities that do not occur in reality. 

% The privacy protection has been drawn increasingly attention due to the tremendous personal data (\eg, selfie image or videos) circulated publicly through social media such as YouTube, FaceBook and Instagram. 
% These vast volume of personal data exacerbates the potential risk of privacy leakage, as they can be used unauthorized in privacy related applications such as face recognition or head pose estimation.

Recently, deep neural network (DNN) based face synthesis has drawn increasing attention. Using personal data (\eg, selfie images or videos) harvested from social network portals (\eg, YouTube, Facebook and Instagram), there are many open-source software tools that one can use to create DeepFakes \cite{fakeapp,DFaker,faceswap-gan,faceswap,DeepFaceLab}, which are synthesized faces of high level of realism.  As DeepFakes challenge the trustworthiness of online media, in recent years, many methods have been developed  \cite{li2018ictu,afchar2018mesonet,nguyen2019multi,li2019exposing,li2020face,li2019hiding,ruiz2020disrupting} to control their potential negative impacts. Much of the effort has been devoted to building DeepFake detectors  \cite{li2018ictu,afchar2018mesonet,li2019exposing,li2020face}. However, given the fast and wide propagation of DeepFake videos on social media, even the best DeepFake detector may not be timely to limit the damage once they emerge online. This can be complemented with methods aiming to break or slow down the process of DeepFakes generation  \cite{li2019hiding,ruiz2020disrupting}, which focus on disrupting the face detection or the synthesis model itself. 

% \sl{add sentence about previous work focus on face detection and neural network step, then the transition to landmark is more natural.}

\begin{figure*}[t]
    \centering
    \includegraphics[width=\linewidth]{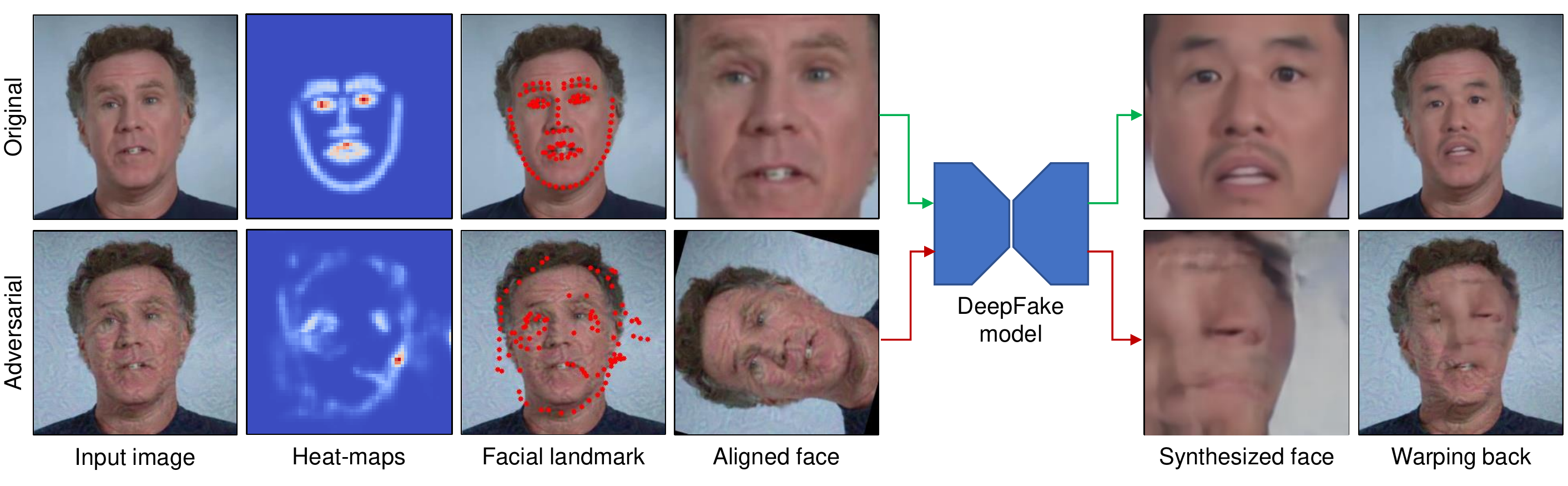}
    \vspace{-0.8cm}
    \caption{\small The overview of Landmark Breaker on obstructing DeepFake generation by disrupting the facial landmark extraction. The top row shows the original DeepFake generation, and the bottom row corresponds to the disruption after facial landmarks are disrupted. The landmark extractor we use is FAN \cite{bulat2017far} and the ``Heat-maps'' is visualized by summing all heat-maps. Note that training of the DeepFake generation model is also affected by disrupted facial landmarks, but is not shown here.}
    \label{fig:overview}
    \vspace{-0.6cm}
\end{figure*}

In this paper, we describe a white-box method to obstruct the creation of DeepFakes based on disrupting the facial landmark extraction, \ie, Landmark Breaker. The facial landmarks are key locations of important facial parts including tips and middle points of eyes, nose, mouth, eye brows as well as contours, see Fig.\ref{fig:overview}. Landmark Breaker attacks the facial landmark extractors by adding adversarial perturbations \cite{szegedy2014intriguing,goodfellow2014explaining}, which are image noises purposely designed to mislead DNN-based facial landmark extractors. Specifically, Landmark Breaker attacks facial landmark heat-maps prediction, which is the common first step in many recent DNN-based facial landmark extractors \cite{bulat2017far,sun2019deep,qian2019aggregation}. We introduce a new loss function to encourage errors between the predicted and original heat-maps to change the final locations of facial landmarks. Then we optimize this loss function using the momentum iterative fast gradient sign method (MI-FGSM) \cite{dong2018boosting}. 

Training the DNN based DeepFake generation model predicates on aligned input faces as training data, which are obtained by matching the facial landmarks of input face to a standard configuration. Also, in the synthesis process of DeepFakes, the facial landmarks are needed to align the input faces. As Landmark Breaker disrupts the essential face alignment step, it can effectively degrade the quality of the DeepFakes, Fig.\ref{fig:overview}. 

We conduct experiments to test Landmark Breaker on attacking three state-of-the-art facial landmark extractors (FAN \cite{bulat2017far}, HRNet \cite{sun2019deep} and AVS-SAN \cite{qian2019aggregation}) using the Celeb-DF dataset \cite{li2020celeb}. The experimental results demonstrate the effectiveness of Landmark Breaker in disrupting the facial landmark extraction as well as obstructing the DeepFake generation. Moreover, we perform ablation studies for different parameter settings and robustness with regards to image and video compression. 

The contribution of this paper is summarized as following:
\begin{enumerate}
    \item[$\bullet$] We propose a new method to obstruct DeepFake generation by disrupting facial landmark extraction. To the best of our knowledge, this is the first study on the vulnerabilities of facial landmark extractors, as well as their application to the obstruction of DeepFake generation.
    \item[$\bullet$] Landmark Breaker is based on a new loss function to encourage the error between predicted and original heat-maps and optimize it using momentum iterative fast gradient sign method.
    \item[$\bullet$] We conduct experiments on three state-of-the-art facial landmark extractors and  study the performance under different settings including video compression. 
\end{enumerate}

\section{Background}
\subsection{DeepFakes}

DeepFake \cite{fakeapp,DFaker,faceswap-gan,faceswap,DeepFaceLab} is a technology recently developed  to create realistic face-swapping videos.
The pipeline of DeepFake synthesis is as follows: (1) Given an input video, the faces of target subject are first  detected using a face detector (\eg, Dlib \cite{dlib09}), and then the facial landmarks are further extracted. (2) The target face is aligned to a standard configuration using the extracted facial landmarks. (3) The aligned faces are then cropped and fed into the DeepFake neural network model to synthesize faces of the donor subject. 
The DeepFake model is based on the auto-encoder model \cite{kingma2013auto} composed by two convolutional neural networks (CNNs), known as the encoder and the decoder, respectively. The encoder aims to remove the identity-related attributes and preserves facial expression and orientation. The decoder reconstructs the source's identity based on the feature obtained from encoder. (4) The synthesized faces are then warped back to the original image and smoothed using masking process that created based on facial landmarks. Therefore, the facial landmarks extraction is an important step to ensure the input of DeepFake model spatially unified. Similarly in training phase, facial landmarks are used as well to align the training faces.

\subsection{Facial landmark extractors}

The facial landmark extractors detect and locate key points of important facial parts such as the tips of the nose, eyes, eyebrows, mouth, and jaw outline. Earlier facial landmark extractors are based on simple machine learning methods such as the ensemble of regression trees (ERT) \cite{kazemi2014one} as in the Dlib package \cite{dlib09}. The more recent ones are based on CNN models, which have achieved significant improved performance over the traditional methods, \eg, \cite{bulat2017far,hu2018facial,wayne2018lab,zou2019learning,sun2019deep,qian2019aggregation}. 
The current CNN based facial landmark extractors typically contain two stages of operations. In the first stage, a set of heat-maps (feature maps) are obtained to represent the spatial probability of each landmark. In the second stage, the final locations of facial landmarks are extracted based on the peaks of the heat-maps. In this work, we mainly focus on attacking the CNN based facial landmark extractors because of their better performance. 

\subsection{Adversarial Perturbations}
CNNs have been proven vulnerable against adversarial perturbations, which are intentionally crafted imperceptible noises aiming to mislead the CNN-based image classifiers  \cite{szegedy2014intriguing,goodfellow2014explaining,kurakin2016adversarial,papernot2016limitations,moosavi2016deepfool,moosavi2017universal,luo2018towards,baluja2018learning,xie2019improving,zeng2019adversarial}, object detectors \cite{xie2017adversarial,chen2018robust}, and semantic segmentation \cite{fischer2017adversarial,arnab2018robustness}.
% \sl{need to introduce white-box/black-box here} 
There are two attack settings: white-box attack, where the attackers can access the details of CNNs, and black-box attack, where the attackers do not know the details of CNNs.
However, to date, there is no existing work to attack CNN-based facial landmark extractors using adversarial perturbations. Compared to the attack to image CNN-based classifiers, which aims to change the prediction of a single label, disturbing facial landmark extractors is more challenging as we need to simultaneously perturb the spatial probabilities of multiple facial landmarks to make the attack effective.

\section{Methodology}

% In this section, we describe in detail of Landmark Breaker. 

\subsection{Notation and Formulation}

Let $\F$ denote the mapping function of a CNN-based landmark extractor of which the parameters we have access to, and $\{h_1,\cdots,h_k\} = \F(\I)$ be the set of heat-maps of running $\F$ on input image $\I$. Our goal is to find an image $\I^{adv}$, which can lead the prediction of landmark locations to a large error, while visually similar to as original image $\I$. The difference $\I^{adv} - \I$ is the adversarial perturbation. We denote the heat-maps from the perturbed image $\{\hat{h}_1,\cdots,\hat{h}_k\} = \F(\I^{adv})$

To this end, we introduce a loss function that aims to enlarge the error between predicted heat-maps and original heat-maps, while constrains the pixel distortion in a certain budget as
\begin{equation}
\begin{array}{cc}
     &  \text{argmin}_{\I^{adv}} L(\I^{adv}, \I) = \sum_{i=1}^k \frac{h_i^\top \hat{h}_i}{\|h_i\|\|\hat{h}_i\|}, \\
     &   s.t. \; || \I^{adv} - \I ||_{\infty} \leq \epsilon,
\end{array}
\label{eq:1}
\end{equation}
where $\epsilon$ is a constant. we use cosine distance to measure the error as it can naturally normalize the loss range in $[-1, 1]$.
% \sl{you have to justify why use the cosine loss, not say L2 or L1 loss.} 
Minimizing this loss function increases the error between predicted and original heat-maps, which will disrupt the facial landmark locations.  

\subsection{Optimization}

We use gradient MI-FGSM \cite{dong2018boosting} method to optimize problem Eq.\eqref{eq:1}. 
% Compared to other methods \cite{goodfellow2014explaining,szegedy2014intriguing}, MI-FGSM integrates the momentum into gradient descent.
% \sl{explain why use MI-FGSM instead of others.}. 
Specifically,
let $t$ denote the iteration number and $\I^{adv}_{t}$ denote the adversarial image obtained at iteration $t$. The start image is initialized as $\I^{adv}_{0} = \I$. $\I^{adv}_{t+1}$ is obtained by considering the momentum and gradient as  
\begin{equation}
\begin{array}{cc}
     &   m_{t+1} = \lambda \cdot m_t + \frac{\nabla_{\I^{adv}} (L(\I^{adv}_t, \I))}{||\nabla_{\I^{adv}} (L(\I^{adv}_t, \I))||_1}, \\
     &  \I^{adv}_{t+1} = {\tt clip} \{ \I^{adv}_{t} - \alpha \cdot {\tt sign}(m_{t+1}) \},
\end{array}
\label{eq:5}
\end{equation}
where $\nabla_{\I^{adv}} (L(\I^{adv}_t,\I))$ is the gradient of $L$ with respect to the input image $\I^{adv}_t$ at iteration $t$; $m_t$ is the accumulated gradient and $\lambda$ is the decay factor of momentum;
$\alpha$ is the step size and {\tt sign} returns the signs of each component of the input vector; {\tt clip} is the truncation function to ensure the pixel value of the resulting image is in $[0, 255]$.  The algorithm stops when the maximum number of iteration $T$ is reached or the distortion threshold $\epsilon$ is reached. The overall algorithm is given in Algorithm \ref{alg:overview}.

\begin{algorithm}[t]
\caption{\footnotesize \em Overview of Landmark Breaker}
\label{alg:overview}
\footnotesize{
	\begin{algorithmic}[1]
		\Require {landmark extractor ${\F}$; input image $\I$; perturbed image $\I^{adv}$; maximal iteration number $T$}
		\State ${\I}^{adv}_0 = {\I}, t = 0, m_0 = 0$
		\While{$t \leq T \;\text{and} \; ||{\I}^{adv}_{t+1} - \I||_{\infty} \leq \epsilon$}
		\State $m_{t+1} = \lambda \cdot m_t + \frac{\nabla_{\I^{adv}} (L(\I^{adv}_t, \I))}{||\nabla_{\I^{adv}} (L(\I^{adv}_t, \I))||_1},$
		\State $\I^{adv}_{t+1} = {\tt clip} \{ \I^{adv}_{t} - \alpha \cdot {\tt sign}(m_{t+1}) \}$
		\State $t = t + 1$
		\EndWhile
		\Ensure Adversarial perturbed image ${\I}^{adv}_t$
    \end{algorithmic}}
\end{algorithm}

%%%%%%%%%%%%%%%%%%%%%%%%%%%%%%%%%%%
%%%%%%%%%%%%%%%%%%%%%%%%%%%%%%%%%%%
\section{Experiments}

\subsection{Experimental Settings}

\begin{figure}[t]
    \centering
    \includegraphics[width=\linewidth]{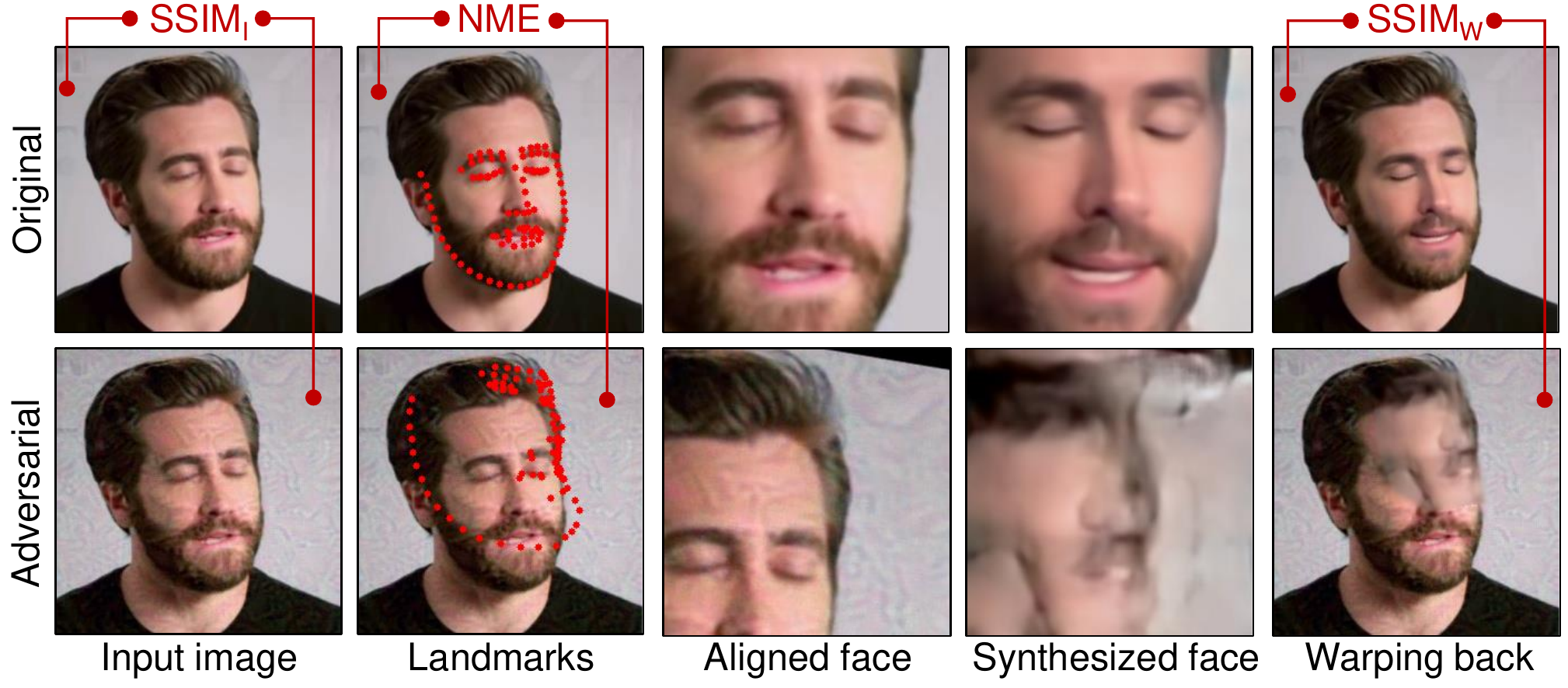}
    \vspace{-0.7cm}
    \caption{\small Evaluation pipeline. SSIM$_I$ denotes the image quality of adversarial image referred to original image, while SSIM$_W$ denotes the image quality of corresponding synthesized image. NME denotes the distance of facial landmarks on adversarial image and ground truth.}
    \label{fig:eval}
    \vspace{-0.4cm}
\end{figure}

\subsubsection{Landmark Extractors} Landmark Breaker is validated on three state-of-the-art CNN based facial landmark extractors, namely, FAN \cite{bulat2017far}, HRNet \cite{sun2019deep} and AVS-SAN \cite{qian2019aggregation}. 
FAN\footnote{\url{https://github.com/hzh8311/pyhowfar}} is constructed by multiple stacked hourglass  structures, where we use one hourglass structure for simplicity.
HRNet\footnote{\url{https://github.com/HRNet/HRNet-Facial-Landmark-Detection}} is composed by parallel high-to-low resolution sub-networks and repeats the information exchange across multi-resolution sub-networks. 
AVS-SAN\footnote{\url{https://github.com/TheSouthFrog/stylealign}} first disentangles face images to style and structure space, which is then used as augmentation to train the network.
We use implementations of all three methods trained on WLFW dataset \cite{wayne2018lab}

\subsubsection{Datasets} To demonstrate the effectiveness of Landmark Breaker on obstructing DeepFake generation, we conduct experiments on Celeb-DF dataset \cite{li2020celeb}, which contains high-quality DeepFake videos of $59$ celebrities. Each video contains one subject with various head pose and facial expression. We choose this dataset as the pretrained DeepFake models are available to us, which can be used to test our method.

In our experiment, we utilize the DeepFake method described in \cite{li2020celeb} to synthesize fake videos using original and adversarial images respectively. We randomly select $6$ identities, corresponding to $36$ videos in total. Since the adjacent frames in a video show little variations, we apply Landmark Breaker to the key frames of each video, \ie, $600$ frames in total, for evaluation. Since the Celeb-DF dataset does not have the ground truth of facial landmarks, we use the results of HRNet as the ground truth due to its superior performance.

\subsubsection{Evaluations}   
We use two metrics to evaluate Landmark Breaker, namely, Normalized Mean Error (NME) \cite{sun2019deep}  and Structural Similarity (SSIM) \cite{wang2004image}. The relation of these metrics are shown in Fig.\ref{fig:eval}. 

\begin{enumerate}
    \item[-] NME is the average Euclidean distance between landmarks on adversarial image and the ground truth, which is then normalized by the distance between the leftmost key point in left eye and the rightmost key point in right eye. Higher NME score indicates less accurate landmark detection, which is the objective of Landmark Breaker.
    \item[-] The SSIM metric simulates perceptual image quality. We use this indicator to demonstrate Landmark Breaker can affect the visual quality of DeepFake. As shown in Fig. \ref{fig:eval}, we compute SSIM of original and adversarial input images (SSIM$_I$)\footnote{We employ mask-SSIM \cite{ma2017pose} to measure the quality inside a region of interest determined by face detection.} and then compute the SSIM of the synthesized results (SSIM$_W$). The lower score indicates the image quality is degraded. Ideally, the attacking method should have large SSIM$_I$ that the adversarial perturbation does not affect the quality of input image, and small SSIM$_W$ that the synthesis quality is degraded. 
\end{enumerate}

\subsubsection{Baselines} 

To better analyze Landmark Breaker, we adapt other two methods FGSM \cite{szegedy2014intriguing} and I-FGSM \cite{goodfellow2014explaining} from attacking image classifiers to our task. Specifically, the FGSM is a single step optimization method as
$\I^{adv}_{1} = {\tt clip} \{ \I^{adv}_{0} - \alpha \cdot {\tt sign}(\nabla_{\I^{adv}_0} (L(\I^{adv}_0, \I)) \},$
while I-FGSM is an iterative optimization method without considering momentum as
$\I^{adv}_{t+1} = {\tt clip} \{ \I^{adv}_{t} - \alpha \cdot {\tt sign}(\nabla_{\I^{adv}} (L(\I^{adv}_t, \I)) \}.$ The step size $\alpha$ and iteration number $T$ of I-FGSM are set as same in Landmark Breaker. We use these two adapted methods as our baseline methods, which are denoted as {\em Base1} and {\em Base2} respectively.

\subsubsection{Implementation Details} Landmark Breaker is implemented by PyTorch 1.0 \cite{paszke2019pytorch} on Ubuntu 18.04, with one Nvidia 1080ti GPU. Following the previous works \cite{luo2015foveation,xie2019improving}, we set the maximum perturbation budget $\epsilon = 15$. The other parameters in Landmark Breaker is set as following: The maximum iteration number $T = 20$; The step size $\alpha = 1$; The decay factor is set as $\lambda = 0.5$. The code of Landmark Breaker will be publicly available upon the acceptance of our paper.

\begin{table}[t]
    \centering
    \caption{The NME and SSIM scores of Landmark Breaker on different landmark extractors. The landmark extractors shown in leftmost column denotes where the adversarial perturbation is from and the ones shown in the top row denotes which landmark extractors are attacked.}
    \label{tab:NME-LB-SSIM}
    \small
    \vspace{-0.3cm}
        NME$\uparrow$\\
        \vspace{0.1cm}
    % \resizebox{0.8\linewidth}{!}{
    \begin{tabular}{|c||c|c|c|}
        \hline
        Attacks & FAN & HRNet & AVS-SAN \\
        \hline
        \hline
        None & 0.03 & 0.00 & 0.09 \\
        \hline
        \hline
        FAN & 0.87 &0.05 &0.09 \\
        \hline
        HRNet & 0.04 & 0.87 &0.09 \\
        \hline
        AVS-SAN & 0.06 & 0.04 & 0.92 \\
        \hline
        \end{tabular}
        % }  
        \vspace{0.2cm}
        \\
        SSIM
        \\
        \vspace{0.1cm}
        \begin{tabular}{|c||c|c|c|c|}
        \hline
        \multirow{2}{*}{Attacks} & \multirow{2}{*}{SSIM$_{I}\uparrow$} & \multicolumn{3}{c|}{SSIM$_{W}\downarrow$} \\
        \cline{3-5}
        & & FAN & HRNet & AVS-SAN \\
        \hline
        \hline
        FAN & 0.81 & 0.68 & 0.89 & 0.89 \\
        \hline
        HRNet & 0.78 & 0.89 & 0.67 & 0.88 \\
        \hline
        AVS-SAN & 0.78 & 0.87 & 0.87 & 0.69 \\
        \hline
        \end{tabular}     
\end{table}

\begin{table}[t]
    \centering
    \caption{The NME and SSIM performance of different attacking methods.}
    \label{tab:NME_SSIM_uncompressed}
    \small
    % \resizebox{0.8\linewidth}{!}{
    \vspace{-0.4cm}
    NME$\uparrow$
    \\
    \vspace{0.2cm}
    \begin{tabular}{|c||c|c|c|}
        \hline
        Attacks & FAN & HRNet & AVS-SAN \\
        \hline
        \hline
        None & 0.03 & 0.00 & 0.09 \\
        \hline
        \hline
        Base1 & 0.05 & 0.04 & 0.10 \\
        \hline
        Base2 & 0.85 & 0.88 & 0.92 \\
        \hline
        LB & 0.87 & 0.87 & 0.92 \\
        \hline
        \end{tabular}
        % }
        % \multicolumn{4}{c}{\vspace{1cm}}\\
        % \vspace{1cm}
        \vspace{0.2cm}
        \\
        SSIM$_{I}\uparrow$ / SSIM$_{W}\downarrow$
        \vspace{0.2cm}
        \\
        \begin{tabular}{|c||c|c|c|}
        \hline
        Attacks & FAN & HRNet & AVS-SAN \\
        \hline
        \hline
        Base1 & 0.52/0.73 & 0.46/0.71 & 0.49/0.69 \\
        \hline
        Base2 & 0.88/0.71 & 0.88/0.70 & 0.86/0.73 \\
        \hline
        LB & 0.81/0.68 & 0.78/0.67 & 0.78/0.69 \\
        \hline
        \end{tabular}
        \vspace{-0.4cm}
\end{table}

\begin{table}[t]
    \centering
    \caption{The NME and SSIM performance of black-box attack. See text for details.}
    \label{tab:black-attack-NME-SSIM}
    \small
    \vspace{-0.3cm}
    NME$\uparrow$
    \\
    \vspace{0.1cm}
    % \resizebox{0.8\linewidth}{!}{
    \begin{tabular}{|c|l||c|c|c|}
        \hline
        \multicolumn{2}{|c||}{Attacks} & FAN & HRNet & AVS-SAN \\
        \hline
        \hline
        \multicolumn{2}{|c||}{None} & 0.03 & 0.00 & 0.09 \\
        \hline
        \hline
        \multirow{2}{*}{FAN} & LB$_{trans}$ &0.22&0.03&0.09
 \\
                             \cline{2-5}
                             & LB$_{mix}$ & 0.24&0.04&0.09  \\

        \hline
        \multirow{2}{*}{HRNet} & LB$_{trans}$ & 0.04& 0.10 &0.09
  \\
                             \cline{2-5}
                             & LB$_{mix}$ &0.04&0.14&0.09
 \\

        \hline
        \multirow{2}{*}{AVS-SAN} & LB$_{trans}$ &0.04&0.03&0.55
  \\
                             \cline{2-5}
                             & LB$_{mix}$ &0.05&0.03&0.56
  \\
        \hline
        % \multicolumn{2}{|c||}{LB$_{ens}$} & 0.05 & 0.05 & 0.09 \\
        % \hline
        % \multicolumn{2}{|c||}{LB$_{ens^*}$} & 0.06 & 0.05 & 0.09 \\
        % \hline
        \end{tabular}
        % }
        \vspace{0.2cm}
        \\
        SSIM
        \\
        \vspace{0.1cm}
        \resizebox{\linewidth}{!}{
       \begin{tabular}{|c|l||c|c|c|c|}
        \hline
        \multicolumn{2}{|c||}{\multirow{2}{*}{Attacks}} & \multirow{2}{*}{SSIM$_{I}\uparrow$} & \multicolumn{3}{c|}{SSIM$_{W}\downarrow$} \\
        \cline{4-6}
        \multicolumn{2}{|c||}{} & & FAN & HRNet & AVS-SAN \\
        \hline
        \hline
        \multirow{2}{*}{FAN} & LB$_{trans}$ & 0.91&0.88&0.94&0.94
 \\
                             \cline{2-6}
                             & LB$_{mix}$ & 0.90&0.86&0.94&0.93
 \\
%                              \cline{2-6}
%                              & LB$_{ens}$ & 0.82&0.90&0.82&0.73
%  \\
%                              \cline{2-6}
%                              & LB$_{ens^*}$ & 0.81&0.88&0.87&0.75
%  \\
        \hline
        \multirow{2}{*}{HRNet} & LB$_{trans}$ &0.92&0.95&0.94&0.95
 \\
                             \cline{2-6}
                             & LB$_{mix}$ &0.90&0.95&0.91&0.94
 \\
%                              \cline{2-6}
%                              & LB$_{ens}$ &0.81&0.73&0.89&0.72
%  \\
%                              \cline{2-6}
%                              & LB$_{ens^*}$ &0.81&0.83&0.88&0.75
%  \\
        \hline
        \multirow{2}{*}{AVS-SAN} & LB$_{trans}$ &0.89&0.94&0.93&0.82
 \\
                             \cline{2-6}
                             & LB$_{mix}$ & 0.88&0.93&0.93&0.81\\
%                              \cline{2-6}
%                              & LB$_{ens}$ &0.83&0.74&0.80&0.90
% \\
%                              \cline{2-6}
%                              & LB$_{ens^*}$ &0.82&0.85&0.77&0.89
%  \\
        \hline
        \end{tabular}
        }
        \vspace{-0.5cm}
\end{table}

\subsection{Results}
Table \ref{tab:NME-LB-SSIM} shows the NME and SSIM performance of Landmark Breaker. The landmark extractors shown in leftmost column denotes where the adversarial perturbation is from and the ones shown in the top row denotes which landmark extractor is attacked. ``None'' denotes no perturbations are added to image. Landmark Breaker can notably increase the NME score and decrease the SSIM$_W$ score in white-box attack (\eg, the value in the row of ``FAN'' and the column of ``FAN'' 
), which indicates Landmark Breaker can effectively disrupt facial landmarks extraction and subsequently affect the visual quality of the synthesized faces. We also compare Landmark Breaker with two baselines Base1 and Base2 in Table \ref{tab:NME_SSIM_uncompressed}. We can observe the Base1 method merely has effect on the NME performance but can largely degrade the quality of adversarial images compared to Base2 and Landmark Breaker (LB). The Base2 method can also achieve the competitive performance with Landmark Breaker in NME but is slightly degraded in SSIM.  

Following existing works attacking image classifiers,  \cite{szegedy2014intriguing,dong2018boosting}, which achieves the black-box attack by transferring the adversarial perturbations from a known model to an unknown model (transferability), we also test the black-box attack using the adversarial perturbation generated from one landmark extractor to attack other extractors. However, the results show that the adversarial perturbations have merely effect on different extractors. 

As shown in Table \ref{tab:NME-LB-SSIM}, the transferability of Landmark Breaker is weak. To improve the transferability, we employ the strategies commonly used in black-box attack on image classifiers: (1) Input transformation \cite{xie2019improving}: we randomly resize the input image and then padding around with zero at each iteration (denoted as LB$_{trans}$); (2) Attacking mixture \cite{xie2019improving}: we alternatively use Base2 and Landmark Breaker to increase the diversity in optimization (denoted as LB$_{mix}$). Table \ref{tab:black-attack-NME-SSIM} shows the results of black-box attack, which reveals that the strategies effective in attacking image classifiers do not work on attacking landmark extractors. This is probably due to the mechanism of landmark extractors is more complex than image classifiers, as the landmark extractors need to output a series of points instead of labels and only a minority of points shifted does not affect the overall prediction.

\begin{figure*}[t]
    \centering
    % \hspace{-0.3cm}
    \hspace{0.5cm} FAN \hspace{5cm} HRNet \hspace{4.5cm} AVS-SAN\\
    \includegraphics[width=0.16\linewidth]{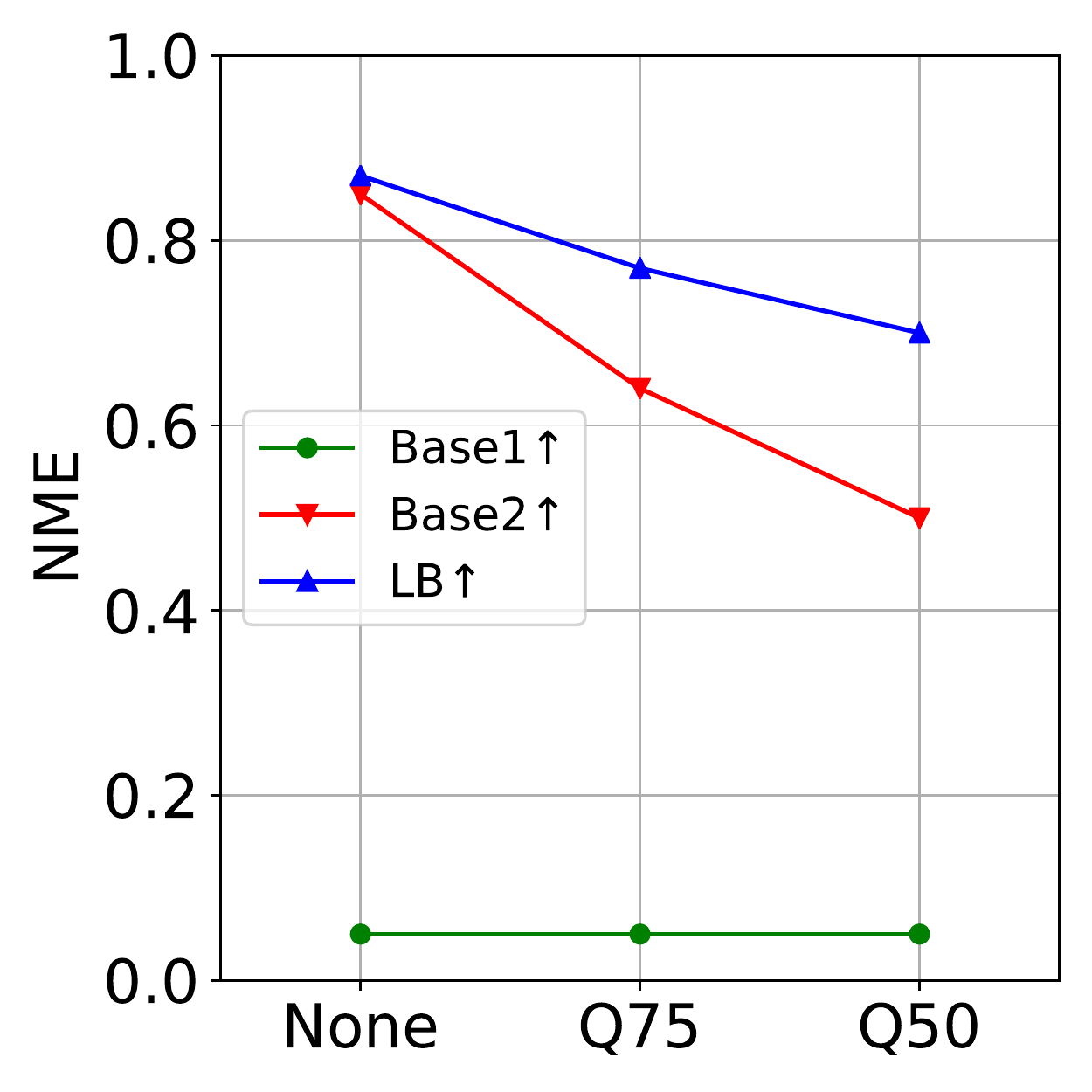}%\hspace{-0.2cm}
    \includegraphics[width=0.16\linewidth]{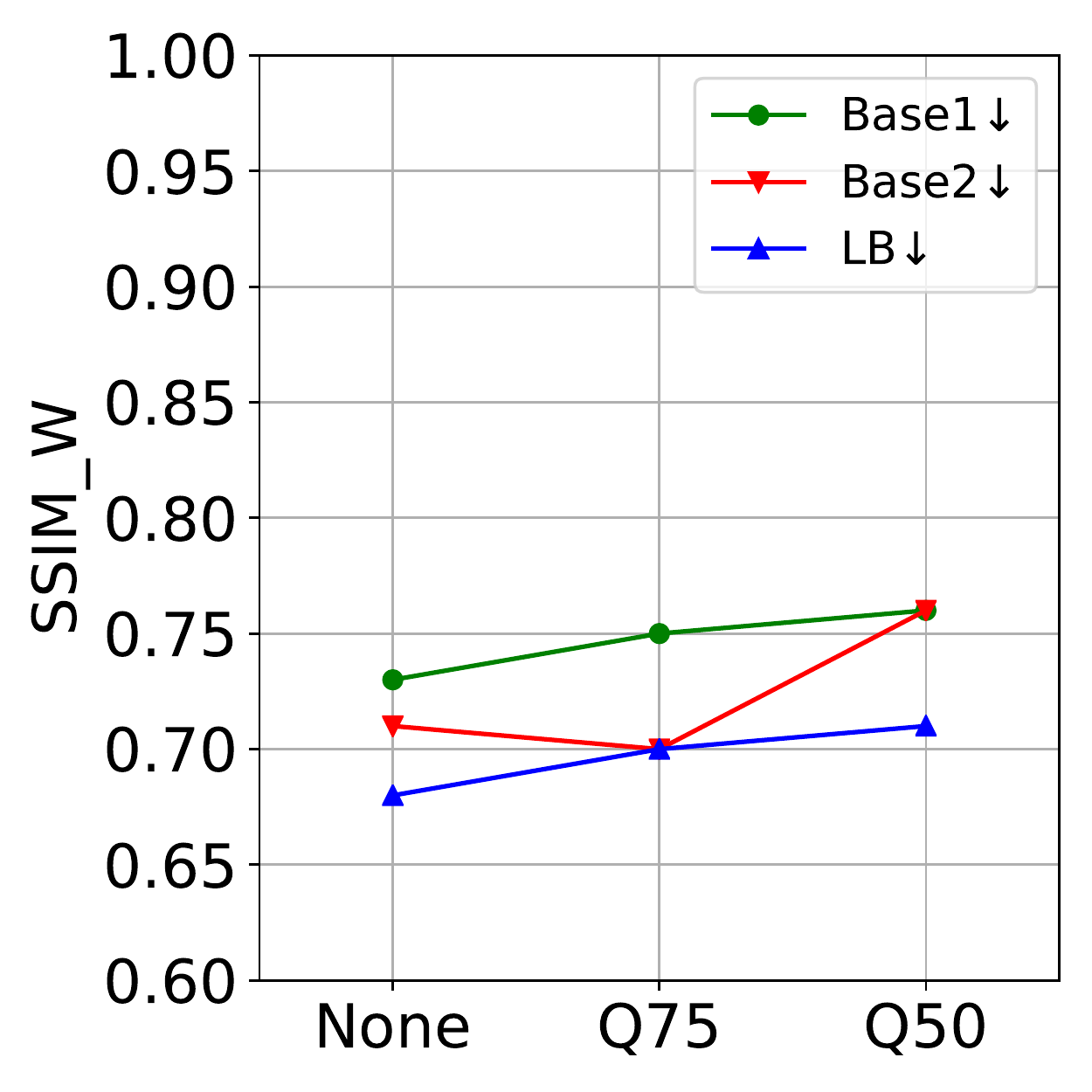}
    \includegraphics[width=0.16\linewidth]{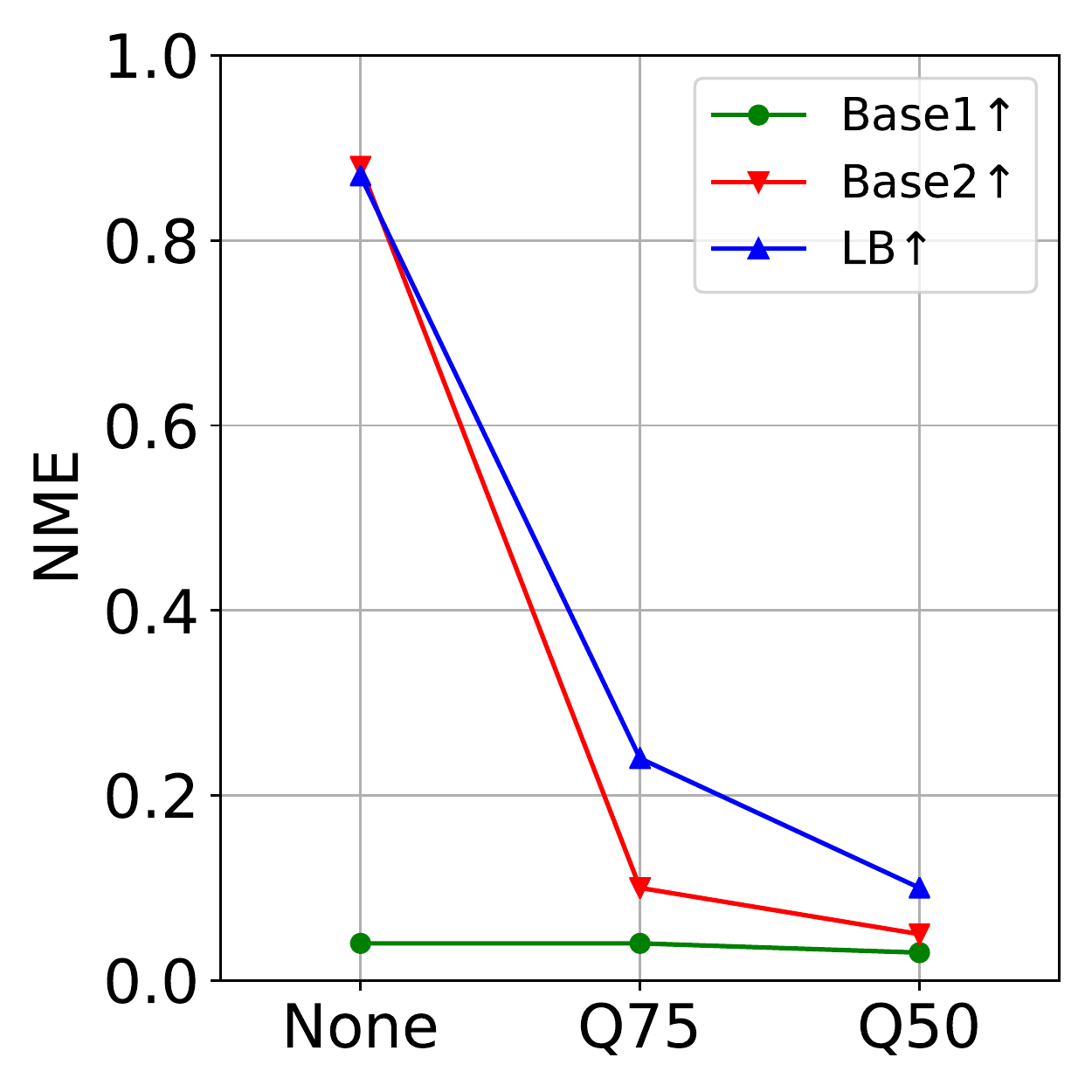}
    \includegraphics[width=0.16\linewidth]{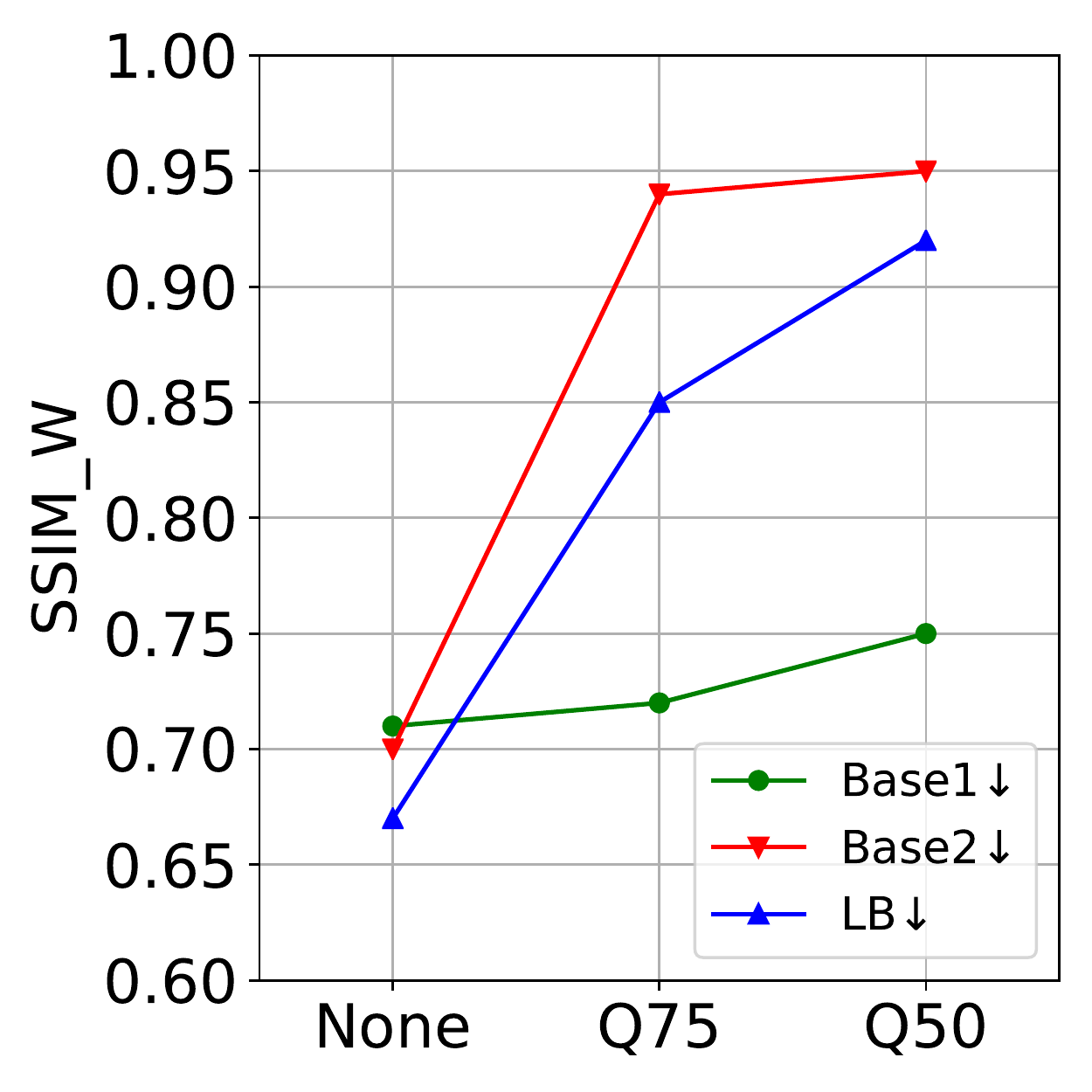}
    \includegraphics[width=0.16\linewidth]{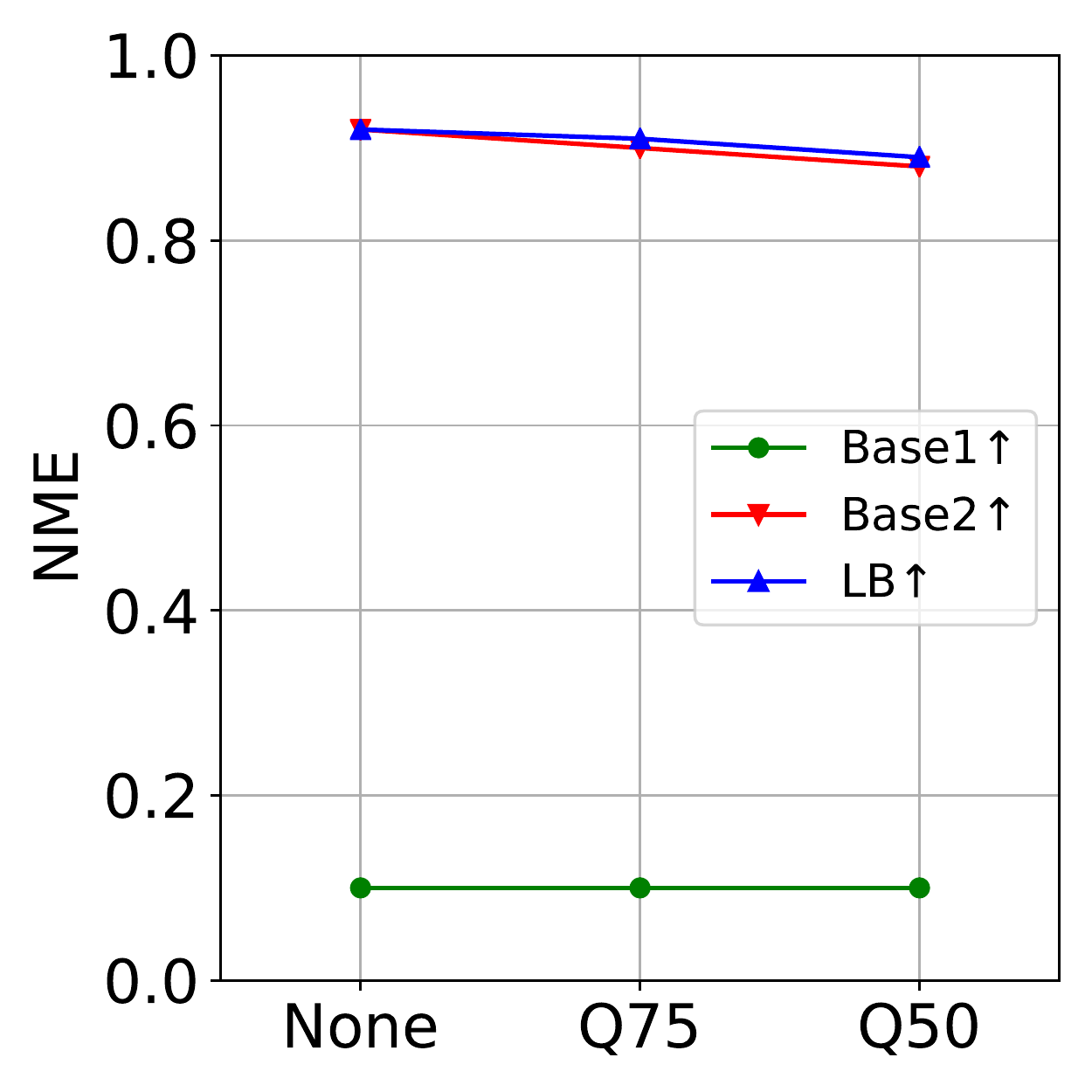}
    \includegraphics[width=0.16\linewidth]{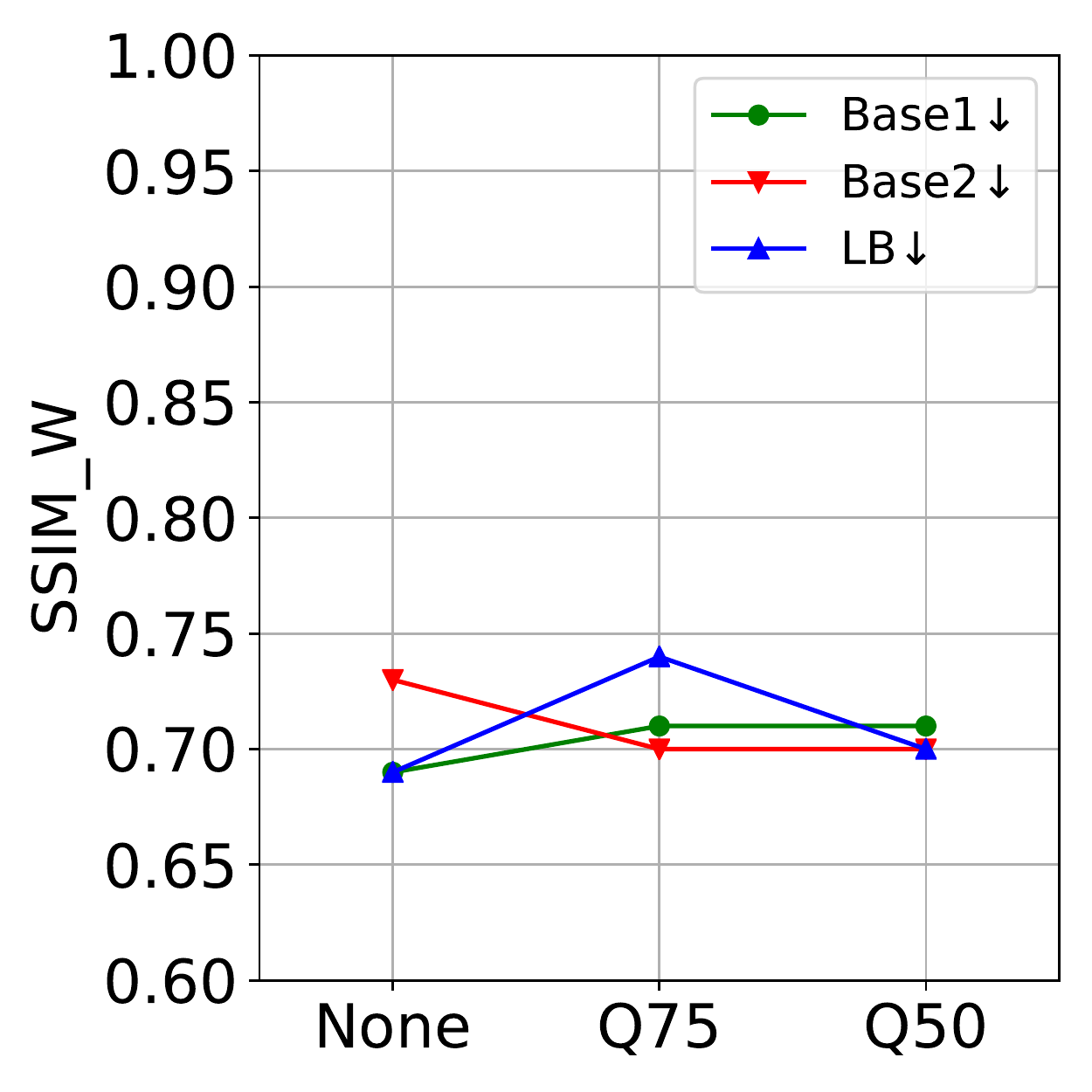}
    
    \hspace{0.01cm}
    \includegraphics[width=0.16\linewidth]{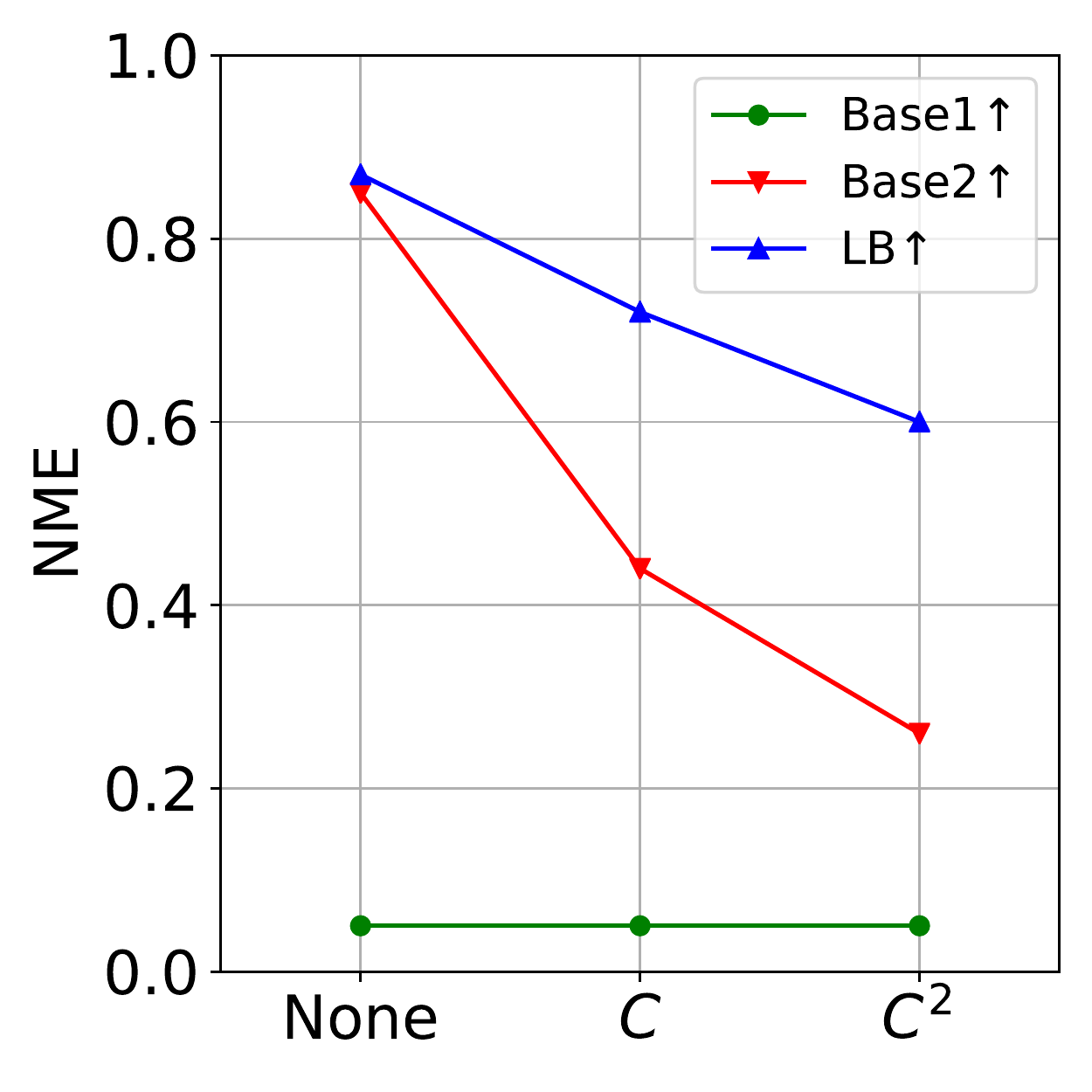}%\hspace{-0.2cm}
    \includegraphics[width=0.16\linewidth]{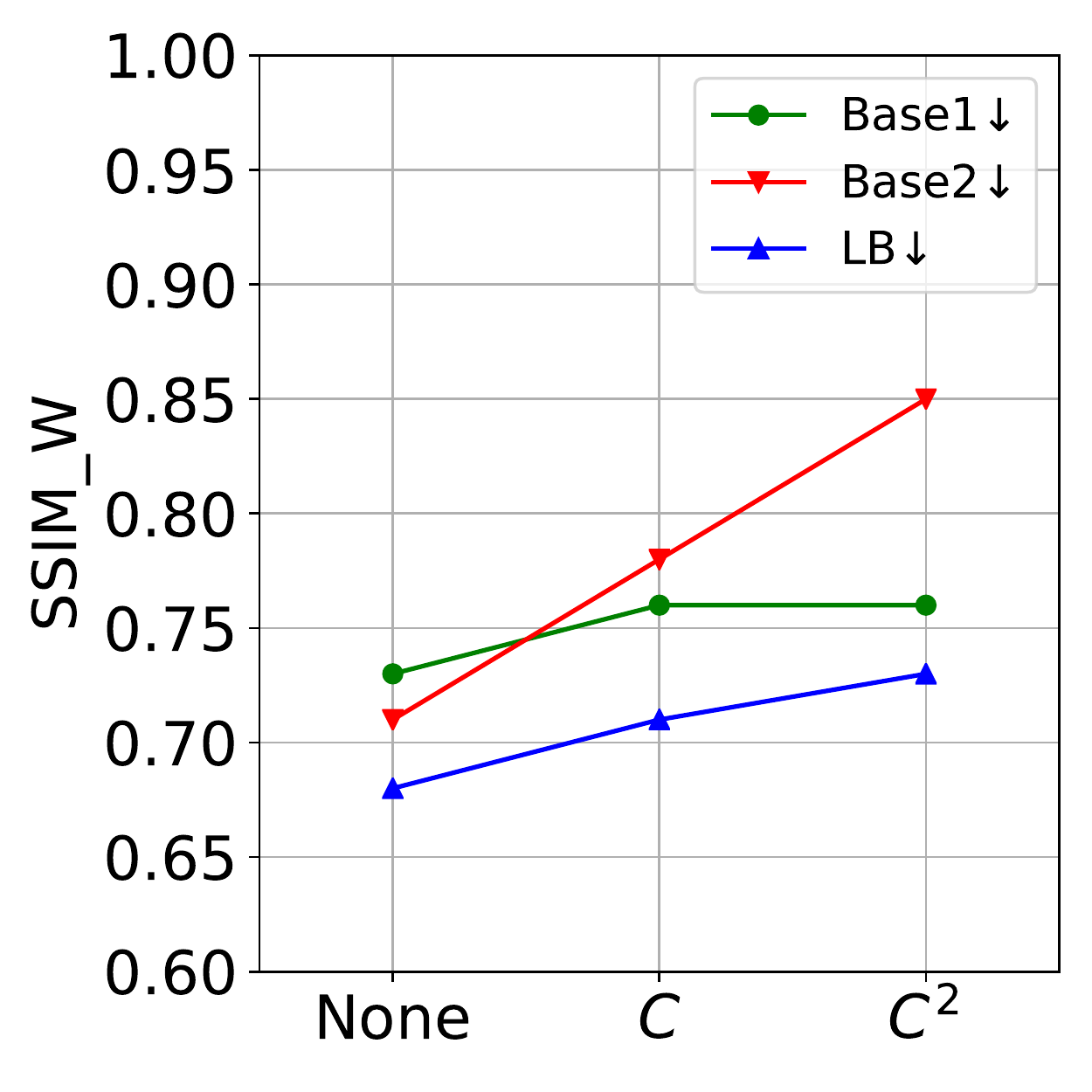}
    \includegraphics[width=0.16\linewidth]{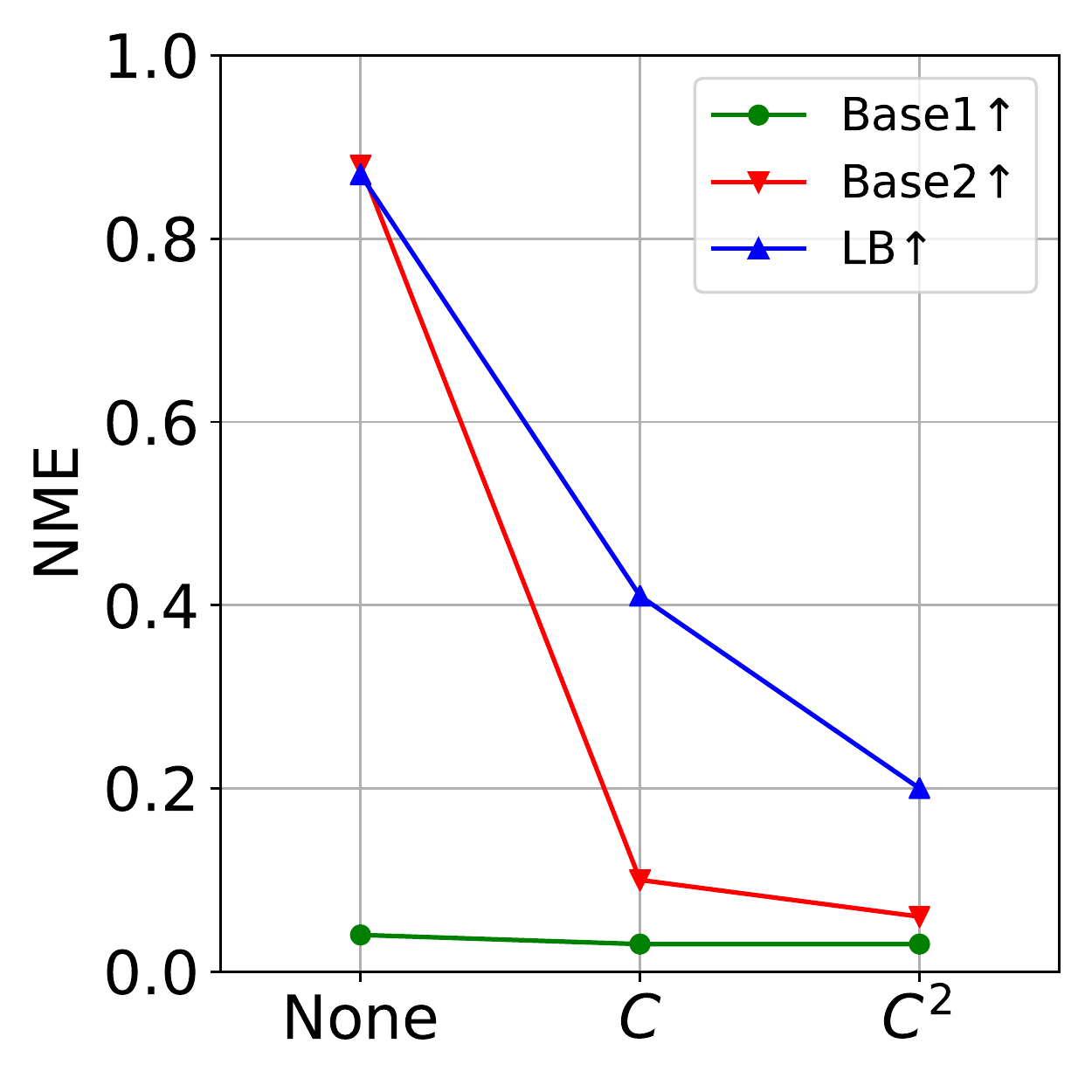}
    \includegraphics[width=0.16\linewidth]{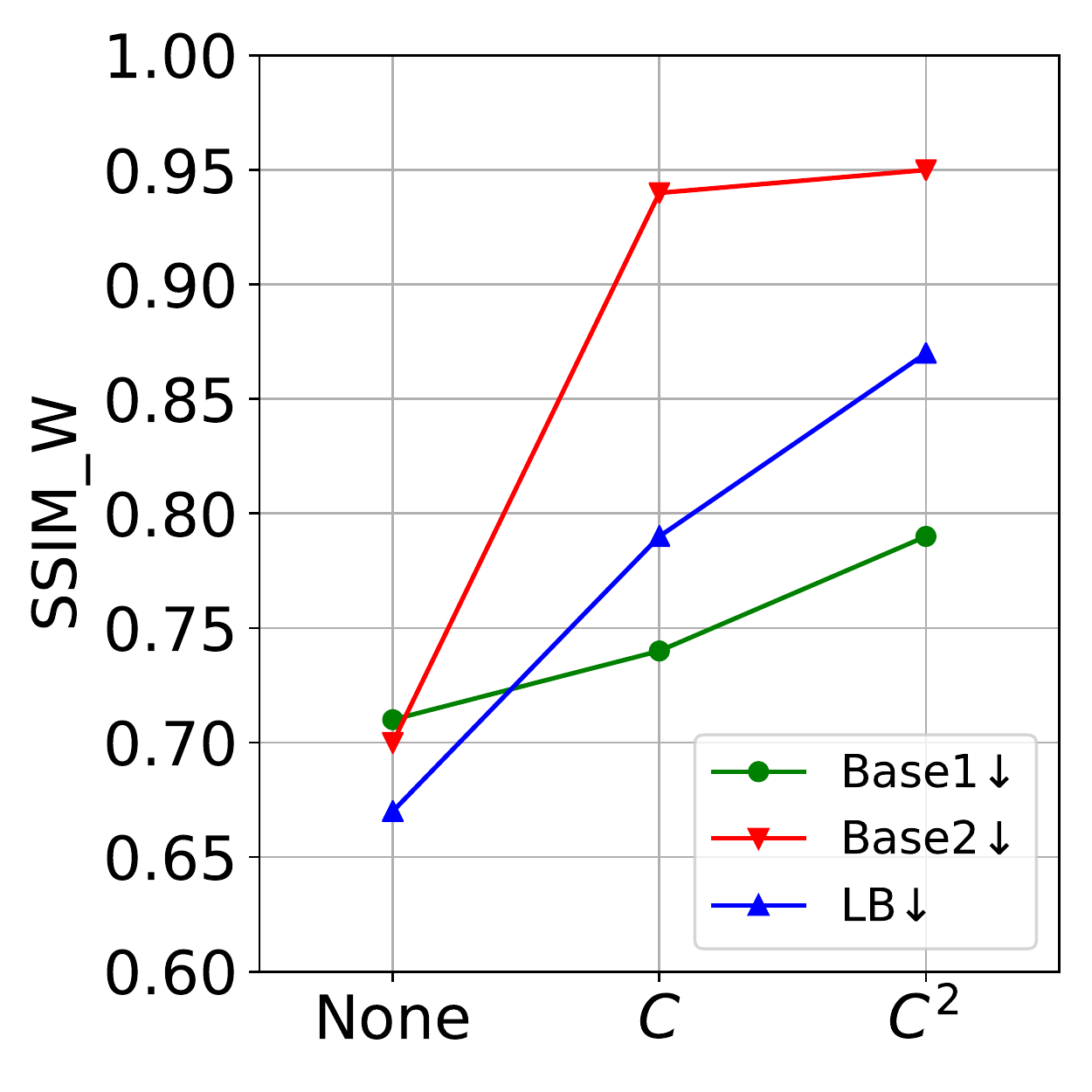}
    \includegraphics[width=0.16\linewidth]{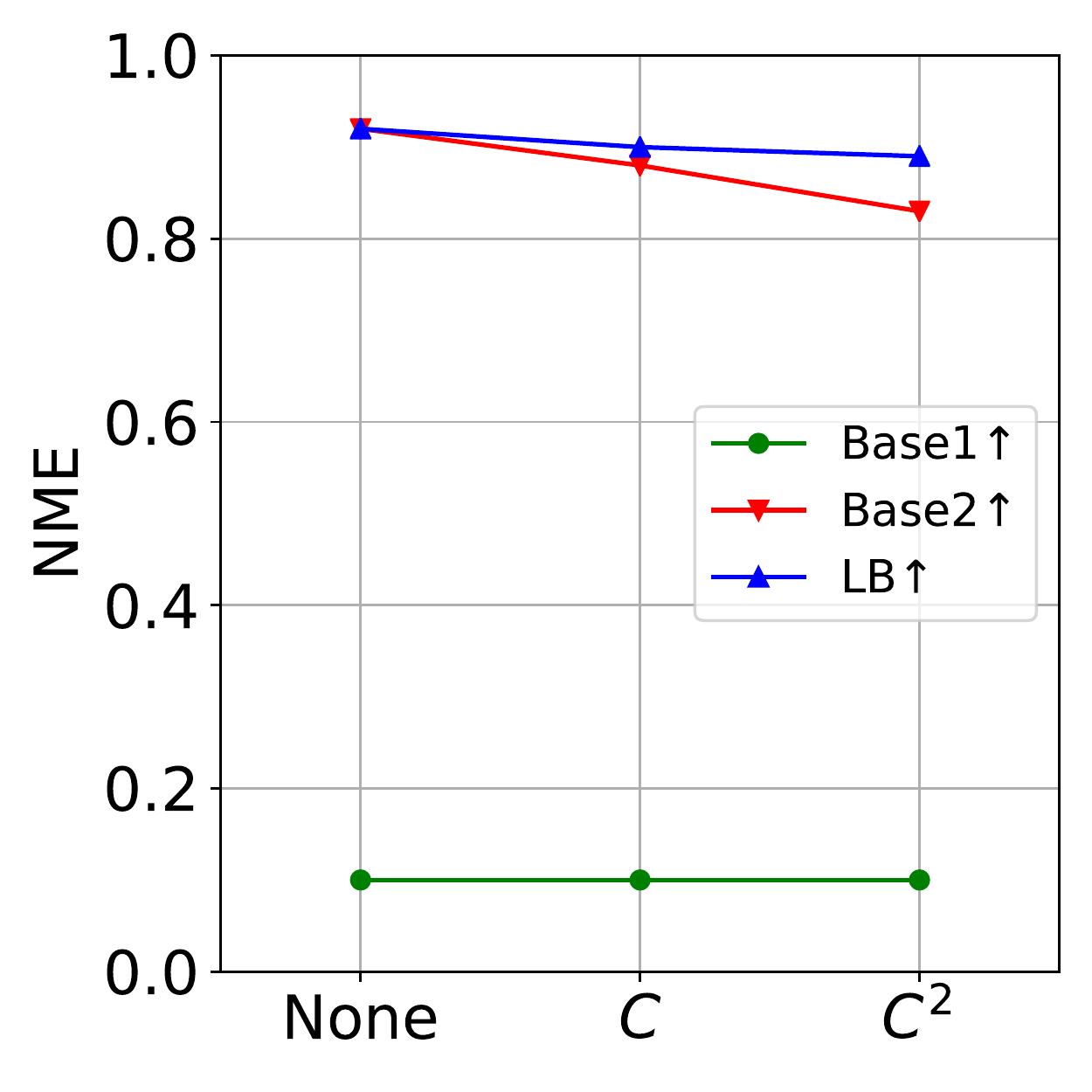}
    \includegraphics[width=0.16\linewidth]{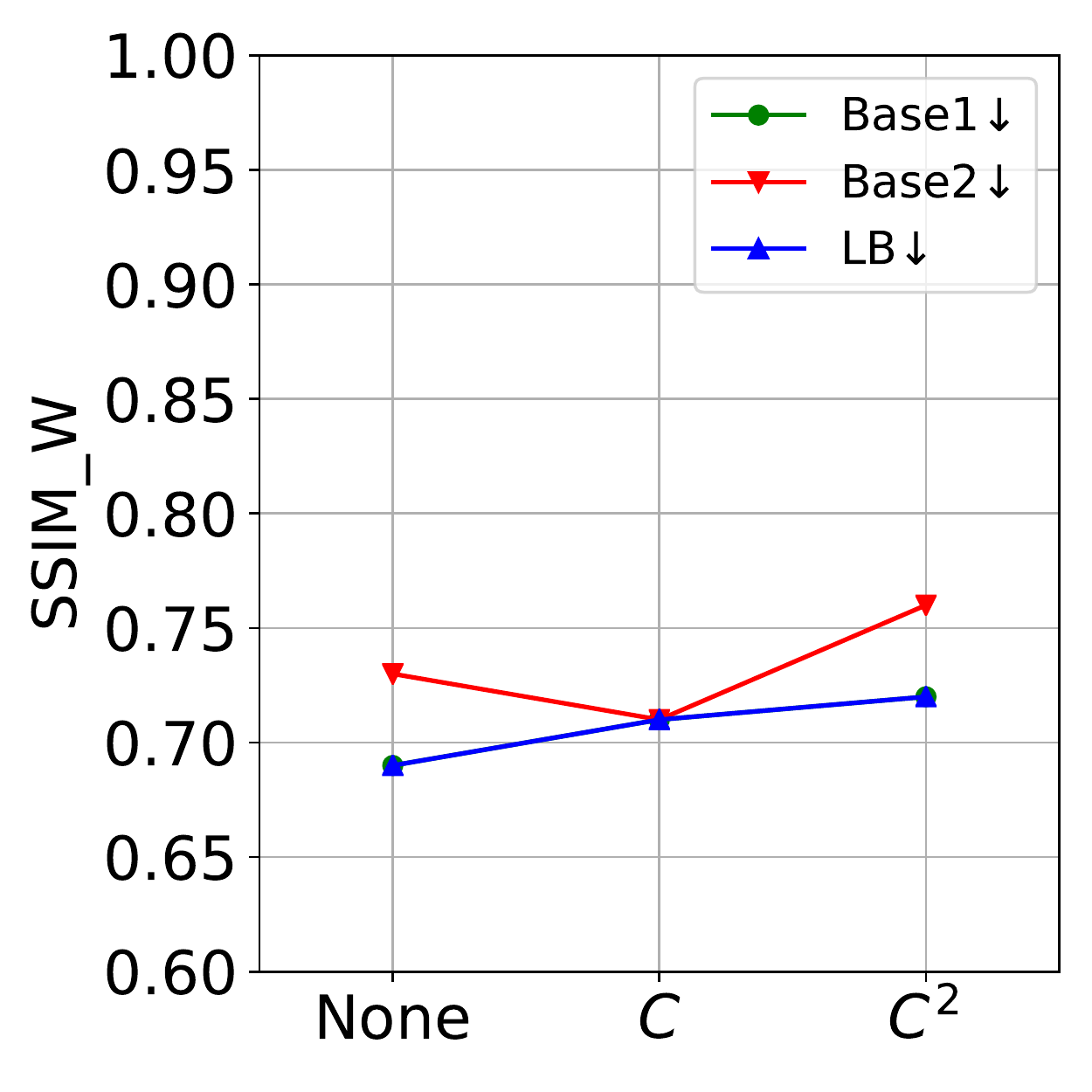}
    \vspace{-0.3cm}
    \caption{\small The performance of each method on different landmark extractors under image and video compression.}
    \label{fig:robust}
    \vspace{-0.4cm}
\end{figure*}

\begin{table}[t]
    \centering
    \caption{The NME and SSIM performance of different attacking methods under different image compression (IC) and video compression (VC) level.}
    \label{tab:comp}
    \small
    \vspace{-0.3cm}
        NME$\uparrow$
        \\
        \vspace{0.1cm}
    % \resizebox{0.8\linewidth}{!}{
    \begin{tabular}{|c|c|l||c|c|c|}
        \hline
        \multicolumn{2}{|c|}{} & Attacks & FAN & HRNet & AVS-SAN \\
        \hline
        \hline
        \multicolumn{2}{|c|}{\multirow{3}{*}{None}} & Base1 &0.05&0.04&0.10
  \\
                             \cline{3-6}
                             \multicolumn{2}{|c|}{} & Base2 &0.85&0.88&0.92\\
                             \cline{3-6}
                             \multicolumn{2}{|c|}{} & LB &0.87&0.87&0.92 \\
        \hline
        \multirow{6}{*}{IC} & \multirow{3}{*}{Q75} & Base1 &0.05&0.04&0.10\\
                             \cline{3-6}
                             & & Base2 &0.64&0.10&0.90\\
                             \cline{3-6}
                             & & LB &0.77&0.24&0.91\\
        \cline{2-6}
        & \multirow{3}{*}{Q50} & Base1 &0.05&0.03&0.10\\
                             \cline{3-6}
                             & & Base2 &0.50&0.05&0.88
 \\
                             \cline{3-6}
                             & & LB &0.70&0.10&0.89\\
        \hline
        \hline
        \multirow{6}{*}{VC} & \multirow{3}{*}{C} & Base1 &0.05&0.03&0.10 \\
                             \cline{3-6}
                             & & Base2 &0.44&0.10&0.88\\
                             \cline{3-6}
                             & & LB &0.72&0.41&0.90\\
        \cline{2-6}
        & \multirow{3}{*}{C$^2$} & Base1 &0.05&0.03&0.10\\
                             \cline{3-6}
                             & & Base2 &0.26&0.06&0.83 \\
                             \cline{3-6}
                             & & LB &0.60&0.20&0.89\\
        \hline
        \end{tabular}
        % }
        \vspace{0.2cm}
        \\
        SSIM$_{I}\uparrow$ / SSIM$_{W}\downarrow$
        \\
        \vspace{0.1cm}
        % \resizebox{\linewidth}{!}{
            \begin{tabular}{|c|c|l||c|c|c|}
        \hline
        \multicolumn{2}{|c|}{}  & Attacks & FAN & HRNet & AVS-SAN \\
        \hline
        \hline
        \multicolumn{2}{|c|}{\multirow{3}{*}{None}} & Base1 &0.52/0.73	&0.46/0.71	&0.49/0.69

 \\
                             \cline{3-6}
                             \multicolumn{2}{|c|}{} & Base2 &0.88/0.71	&0.88/0.70	&0.86/0.73
\\
                             \cline{3-6}
                             \multicolumn{2}{|c|}{} & LB &0.81/0.68	&0.78/0.67	&0.78/0.69
\\
        \hline
        \multirow{6}{*}{IC} &  \multirow{3}{*}{Q75} & Base1 &0.56/0.75	&0.47/0.72	&0.51/0.71
\\
                             \cline{3-6}
                             & & Base2 &0.90/0.74	&0.93/0.94	&0.87/0.74
\\
                             \cline{3-6}
                             & & LB &0.84/0.70	&0.85/0.85	&0.80/0.70
\\
        \cline{2-6}
        &  \multirow{3}{*}{Q50} & Base1 &0.59/0.76	&0.55/0.75	&0.52/0.71
\\
                             \cline{3-6}
                             & & Base2 &0.89/0.76	&0.93/0.95	&0.87/0.74

\\
                             \cline{3-6}
                             & & LB &0.84/0.71	&0.88/0.92	&0.80/0.70
\\
        \hline
        \hline
        \multirow{6}{*}{VC} & \multirow{3}{*}{C} & Base1 &0.57/0.76	&0.52/0.74&0.52/0.71
\\
                             \cline{3-6}
                             & & Base2 &0.90/0.78	&0.92/0.94	&0.87/0.74
\\
                             \cline{3-6}
                             & & LB &0.84/0.71	&0.84/0.79	&0.80/0.71
\\
        \cline{2-6}
        & \multirow{3}{*}{C$^2$} & Base1 &0.58/0.76	&0.52/0.74&0.53/0.72
\\
                             \cline{3-6}
                             & & Base2 &0.91/0.85	&0.93/0.95	&0.88/0.76
\\
                             \cline{3-6}
                             & & LB &0.85/0.73	&0.84/0.87	&0.82/0.72
\\
        \hline
        \end{tabular}
        % }
        \vspace{-0.6cm}
\end{table}

\subsection{Robustness Analysis}
\label{sec:robustness}

We study the robustness of Landmark Breaker towards three extractors under image and video compression. Note image compression considers the spatial correlation, while video compression also considers the temporal correlation. 

\subsubsection{Image compression} We compress the adversarial images to quality $75\%$ (Q75) and $50\%$ (Q50) using OpenCV and then observe the variations in performance of each method. Table \ref{tab:comp} shows the NME and SSIM performance of each method under different compression levels. Compared to the two baseline methods, Landmark Breaker is more robust against image compression. Another observation is that the attacks on AVS-SAN exhibit high robustness, where the performance of NME and SSIM is only slightly degraded. In contrast, the attacking performance on HRNet drops quickly with compression. Fig. \ref{fig:robust} (top) plots the trend of each method. 

\subsubsection{Video compression}  As the videos are widespread on Internet, we also investigate the robustness against video compression. We create a video using the adversarial images using the codec in MPEG4 (denoted as C) and then separate the videos into frames to test the performance.
We also perform double compression to the MPEG4 videos using the codec in H264 (denoted as C$^2$). Table \ref{tab:comp} also shows the performance against video compression, which has the same trend as in image compression. Compared to the baseline methods, Landmark Breaker is more robust. Also, the attacks on AVS-SAN exhibit strong robustness even after double compression C$^2$, on the other hand, the attacks on HRNet are vulnerable against video compression, see Fig. \ref{fig:robust} (bottom). Note the curve of Base1 and LB are fully overlapped in the last plot.

\begin{figure*}[t]
    \centering
    % \hspace{-0.3cm}
    \includegraphics[width=0.16\linewidth]{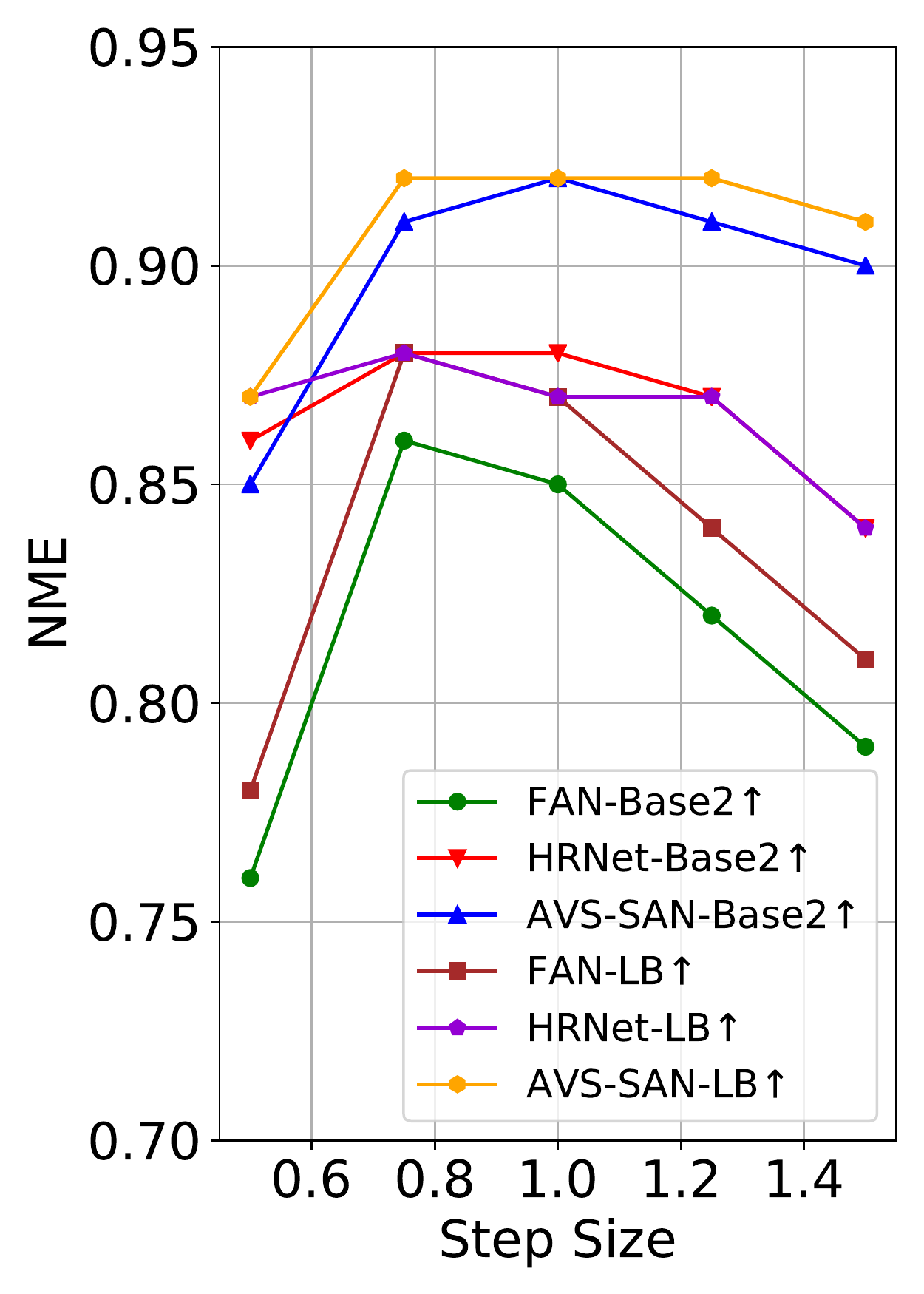}%\hspace{-0.2cm}
    \includegraphics[width=0.16\linewidth]{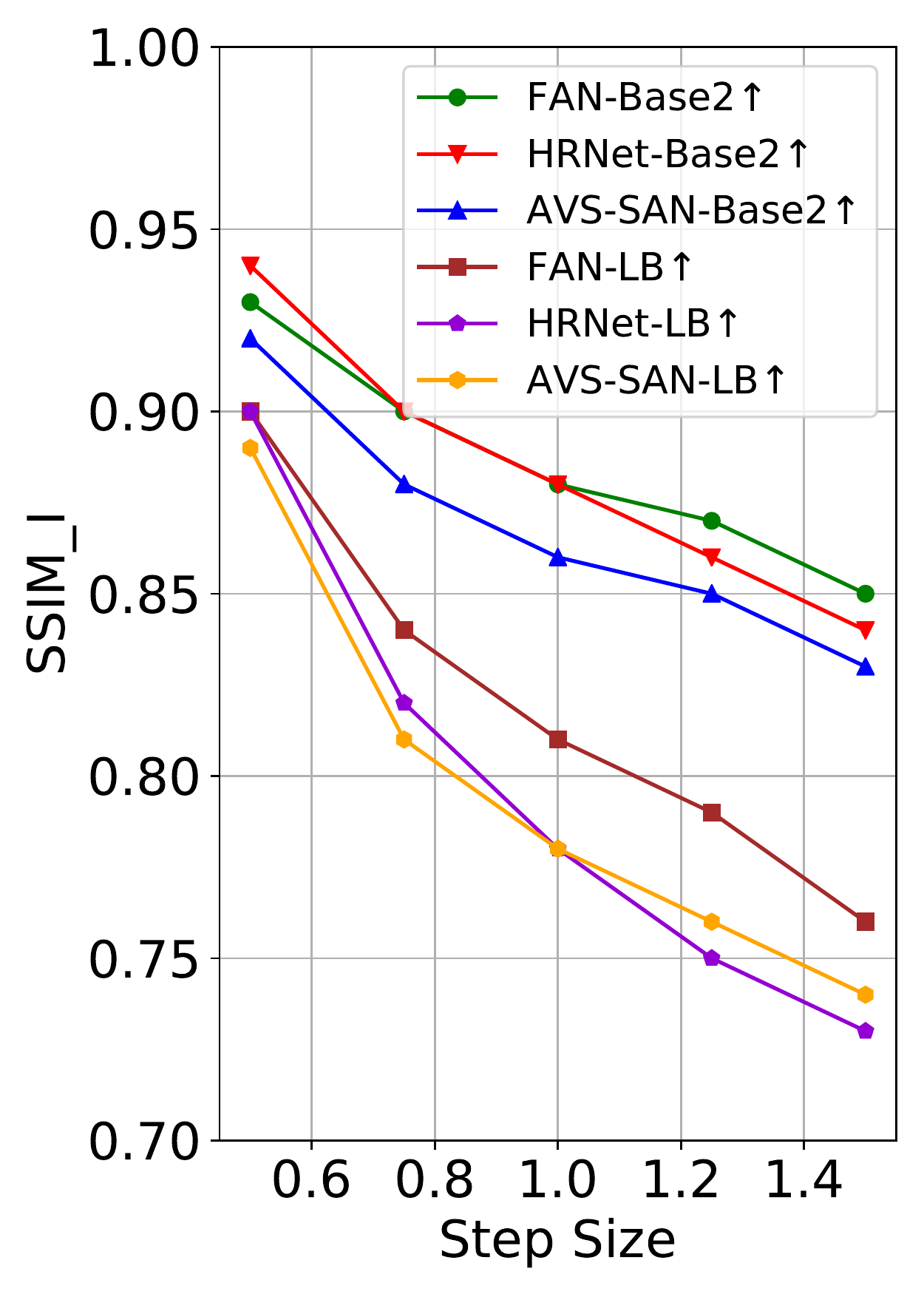}
    \includegraphics[width=0.16\linewidth]{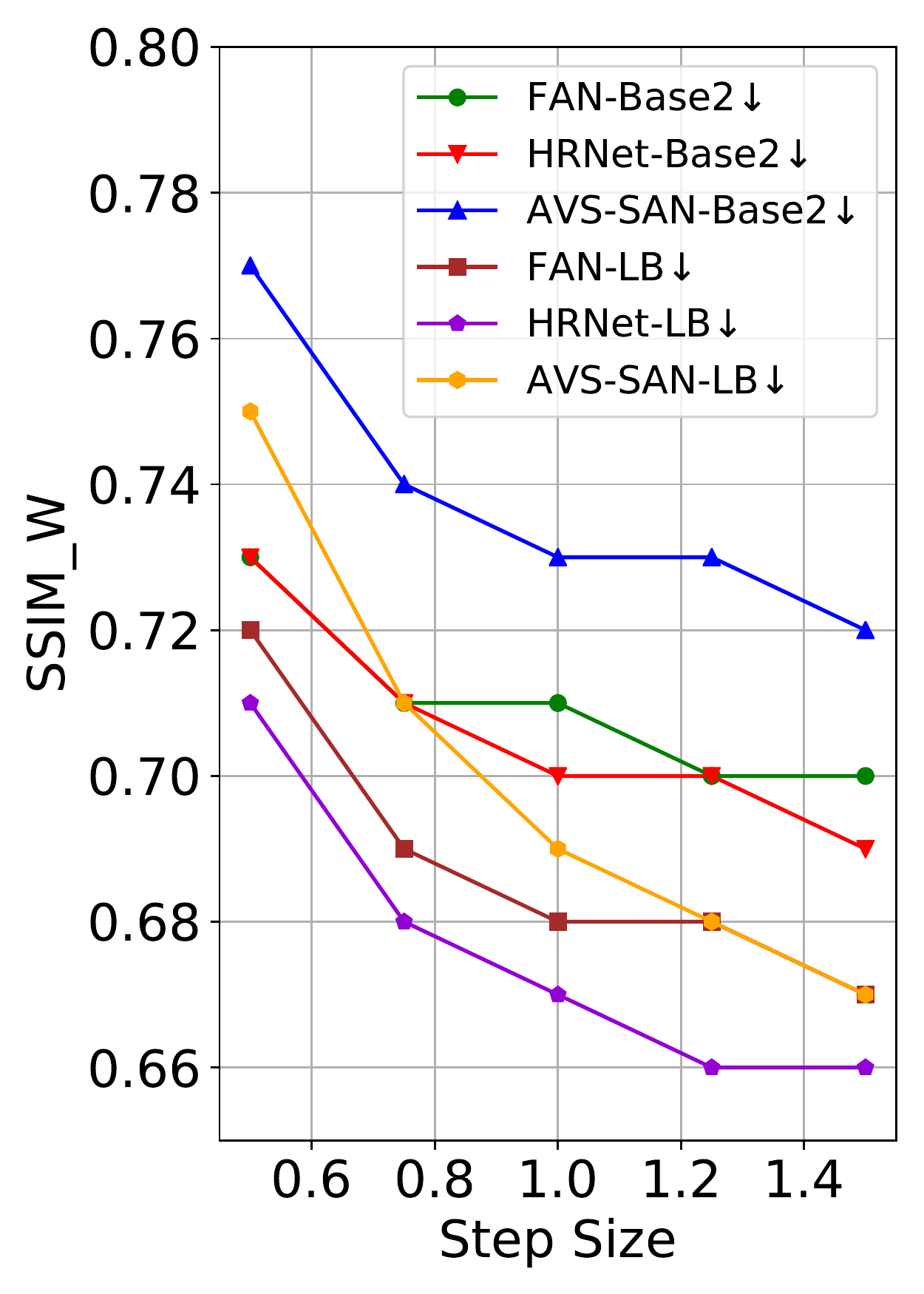}
    \includegraphics[width=0.16\linewidth]{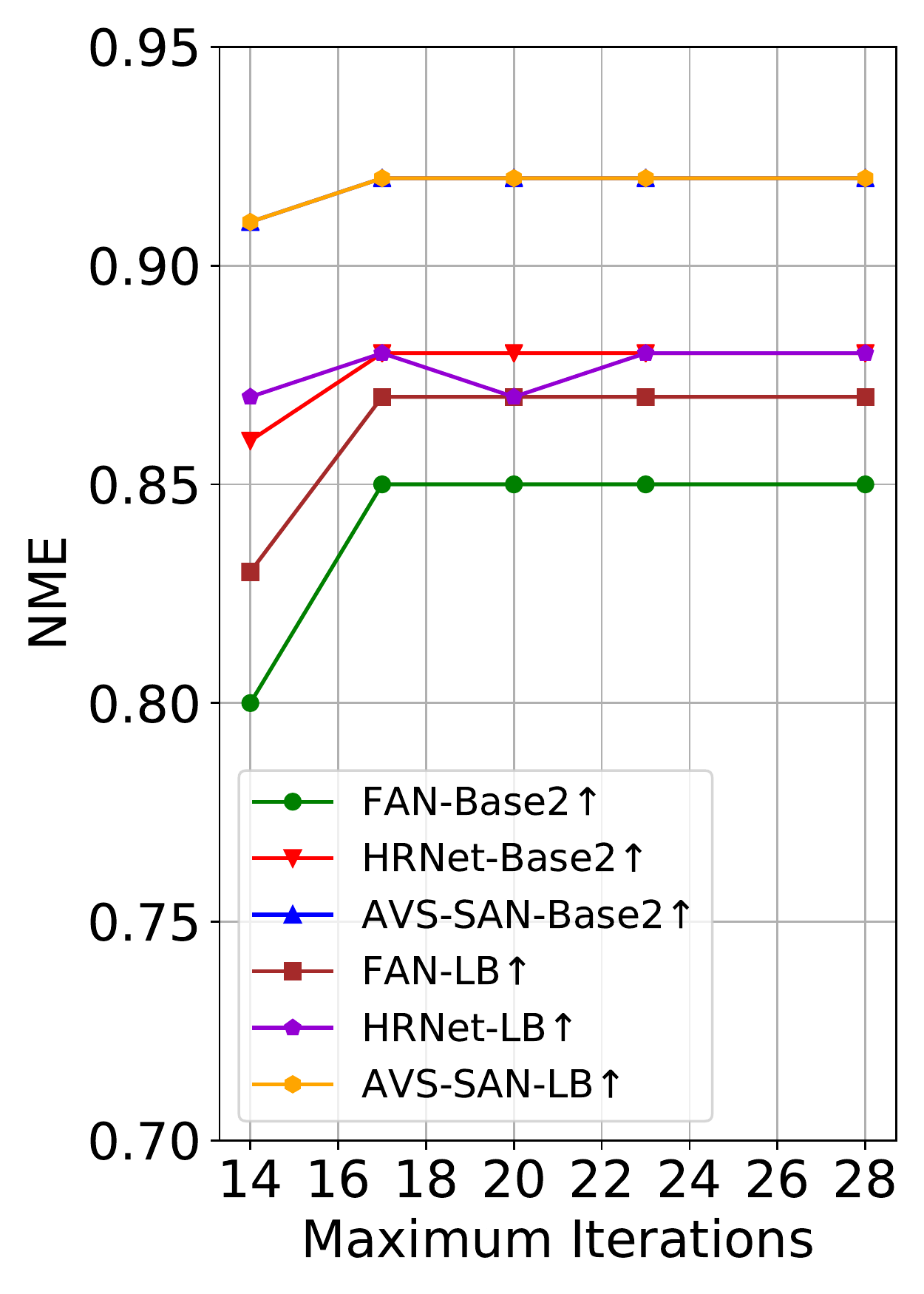}%\hspace{-0.2cm}
    \includegraphics[width=0.16\linewidth]{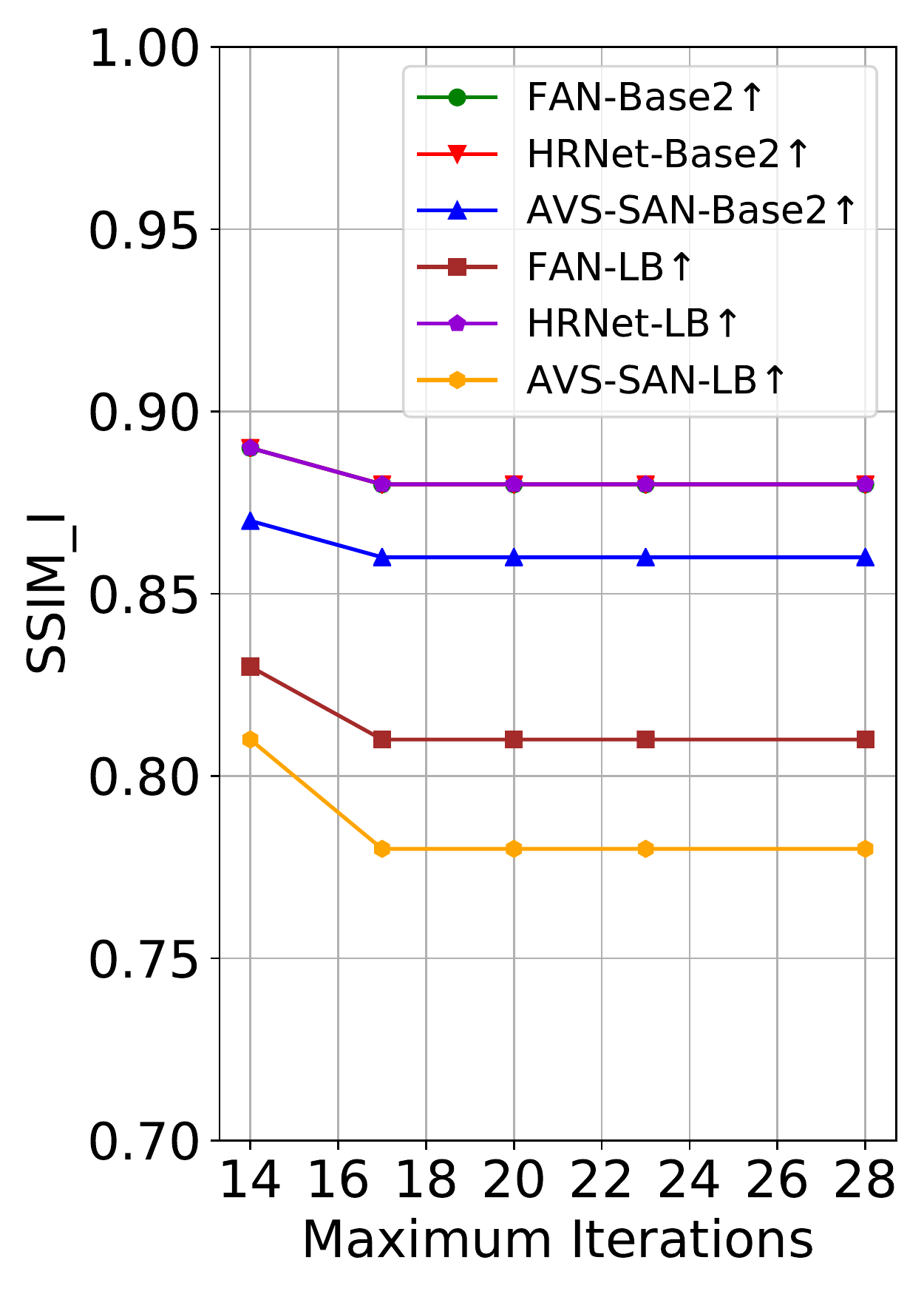}
    \includegraphics[width=0.16\linewidth]{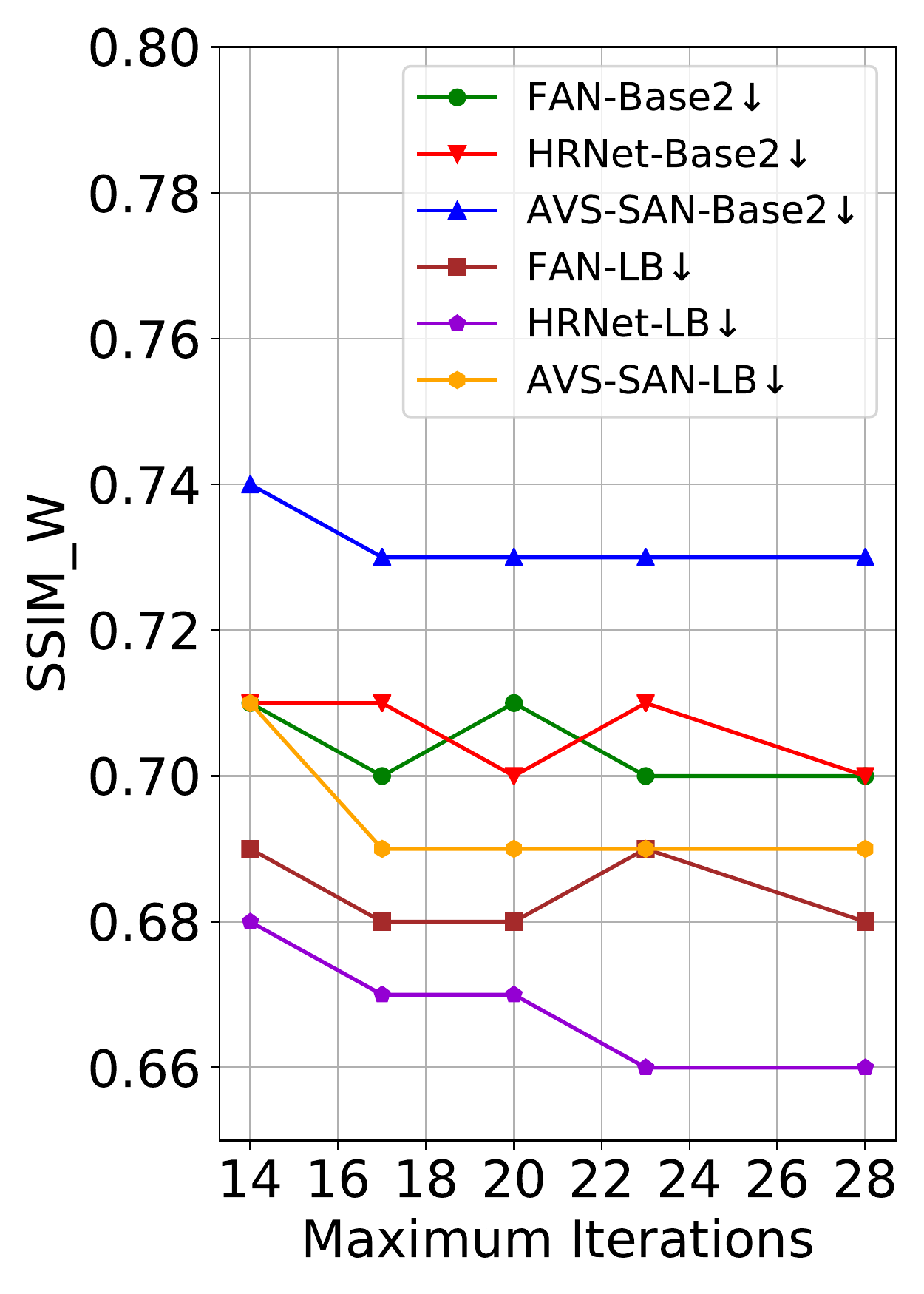}
    \vspace{-0.3cm}
    \caption{\small Ablation study of Landmark Breaker regarding the performance with different step size and iteration number.}
    \label{fig:ablation}
    \vspace{-0.4cm}
\end{figure*}

\subsection{Ablation Study}
This section presents ablation studies on the impact of different parameters to Landmark Breaker.

\subsubsection{Step size} We study the impact of step size $\alpha$ on the performance of NME and SSIM scores. We set the step size $\alpha$ from $0.5$ to $1.5$. The results are plotted in Fig. \ref{fig:ablation}. We observe the NME score increases first and then decreases, which is because the small step size does not disturb the image enough within the maximum iteration number and then the large step size may not precisely follow the gradient descent direction. Moreover, larger step size can degrade the input image quality, which also leads the degradation of the synthesized image. 
\subsubsection{Maximum iteration number} We then study the impact of the maximum iteration number $T$ on the performance of NME and SSIM. We varies the maximum iteration number $T$ from $14$ to $28$ and illustrate the results in Fig. \ref{fig:ablation}. From the figure we observe the NME score is increased and SSIM is decreased with iteration number increasing. Since the distortion budget constraint, the curve becomes flat after about $17$ iterations. Note several curves are fully overlapped in the plot.

\section{Conclusion}
This paper describes a new method, namely Landmark Breaker, to obstruct the DeepFake generation by breaking the prerequisite step -- facial landmark extraction. To do so, we create the adversarial perturbations to disrupt the facial landmark extraction, such that the input faces to the DeepFake model cannot be well aligned. Landmark Breaker is validated on Celeb-DF dataset, which demonstrates the efficacy of Landmark Breaker on disturbing facial landmark extraction. We also study the performance of Landmark Breaker under various parameter settings. 

% Our future work will try to develop effective ways to improve the transferibility of adversarial perturbation for the black-box attack.  

\noindent{\bf Acknowledgement}. This work is partly supported by a US National Science Foundation research grant (IIS-2008532). 

\bibliographystyle{IEEEtran}
% argument is your BibTeX string definitions and bibliography database(s)
\bibliography{ref}

\end{document}